\documentclass[11pt,onecolumn]{article}

\usepackage[utf8]{inputenc} 
\usepackage[english]{babel}
\usepackage{graphicx} 
\usepackage[a4paper,top=25mm,bottom=35mm,left=13mm,right=13mm,]{geometry}

\usepackage[hidelinks, breaklinks]{hyperref} 

\usepackage{amsmath,amssymb,amsfonts} 
\usepackage{diagbox}
\usepackage{multicol}

\usepackage{palatino} 
\usepackage{microtype} 
\usepackage{upgreek} 
\usepackage{csquotes} 
\usepackage{lipsum}
\MakeOuterQuote{"} 

\usepackage{booktabs,multirow} 
\usepackage{caption} 
\usepackage{subcaption}
\usepackage{authblk} 

\usepackage[
    backend=biber,
    style=numeric,
]{biblatex}
\usepackage{fancyhdr}
\usepackage{url}

\usepackage{enumitem}
\setlist[itemize]{topsep=0pt,leftmargin=6mm}
\setlist[enumerate]{topsep=0pt,leftmargin=6mm}

\usepackage{lmodern}
\usepackage{slantsc}

\usepackage{xfrac}

\usepackage[ruled,linesnumbered]{algorithm2e}
\SetKwRepeat{Do}{do}{while}

\usepackage{float}

\usepackage[T1]{fontenc}
\usepackage{newpxtext}
\usepackage[sc]{titlesec}
\titleformat{\section}{\scshape\bfseries\Large}{\thesection}{1em}{}[]

\newcommand\blfootnote[1]{%
  \begingroup
  \renewcommand\thefootnote{}\footnote{#1}%
  \addtocounter{footnote}{-1}%
  \endgroup
}

\graphicspath{ {./figs/} }
\newcommand{\concat}{\mathop{\|}\limits}

\newcommand{\mytitle}{Advanced Graph Clustering Methods: A Comprehensive and In-Depth Analysis}
\title{\huge{\centering \mytitle}}
\author[1]{Timothé Watteau}
\author[1]{Aubin Bonnefoy}
\author[1]{Simon Illouz--Laurent}
\author[1]{Joaquim Jusseau}
\author[1]{Serge Iovleff}
\affil[1]{UTBM, Energy and IT Division, F-90010 Belfort cedex, France}

\date{}
\begin{document}

\begin{center} 
    \parbox{\dimexpr \linewidth-3.5cm}{\maketitle}
\end{center}

\begin{abstract}
Graph clustering, which aims to divide a graph into several homogeneous groups, is a critical area of study with applications that span various fields such as social network analysis, bioinformatics, and image segmentation. This paper explores both traditional and more recent approaches to graph clustering. Firstly, key concepts and definitions in graph theory are introduced. The background section covers essential topics, including graph Laplacians and the integration of Deep Learning in graph analysis. The paper then delves into traditional clustering methods, including Spectral Clustering and the Leiden algorithm. Following this, state-of-the-art clustering techniques that leverage deep learning are examined. A comprehensive comparison of these methods is made through experiments. The paper concludes with a discussion of the practical applications of graph clustering and potential future research directions.
\end{abstract}

\vspace*{-.15\baselineskip}
\begin{center}
    \rule[.8mm]{0.25\textwidth}{.4pt} $\diamond$ \rule[.8mm]{0.25\textwidth}{.4pt}
\end{center}
\vspace*{-.35\baselineskip}

\blfootnote{The implementation used for the experiments is available at \href{https://github.com/timothewt/AdvancedGraphClustering}{https://github.com/timothewt/AdvancedGraphClustering}.}


\begin{multicols}{2}

\section{Introduction}

Cluster Analysis, or clustering, is an unsupervised learning task that aims to partition a set of objects without prior knowledge of their relationships \cite{Everitt_Landau_Leese_Stahl_2011}. Within each group (or cluster), the data points should be as similar as possible, while ensuring that the clusters themselves are as distinct as possible from one another. 

The growing effectiveness of clustering algorithms on straightforward data has sparked interest in extending clustering techniques to more intricate data structures, especially those in the form of graphs \cite{von_Luxburg_2007}. Unlike traditional clustering, which operates on isolated data points, these methods target clustering nodes within graphs. Graphs serve as versatile representations of various real-world systems, including road networks, social networks, web pages, and molecular structures \cite{Newman_2010}. This task is commonly referred to as Graph Clustering or Node Clustering.

Using the connections and arrangement of the nodes within a graph, graph clustering offers unique and powerful insights, which prove to be especially valuable for datasets where inherently relationships are of significant importance, such as social networks, biological networks, and information networks \cite{Fortunato_2010}. In particular, graph clustering has demonstrated superior performance compared to traditional algorithms when applied to tabular data \cite{Shi_Malik_2000}.

In recent times, the surge in interest surrounding Deep Learning and its adeptness in representing non-linear relationships, enabling the discovery of hidden structure and automatic feature extraction has brought Graph Clustering to state-of-the-art performance.

The objective of this paper is to provide a thorough and detailed examination of various graph clustering methodologies, including both traditional approaches and contemporary methods that leverage the capabilities of deep learning \cite{Bengio_Courville_Vincent_2012}.

This study is organized as follows: Section \ref{sec:background} introduces the fundamental definitions in graph theory and the basics of graph clustering, and also covers essential background concepts, including graph Laplacians and an overview of Deep Learning. Traditional clustering methods, such as Spectral Clustering and the Leiden algorithm, are discussed in Section \ref{sec:tradmethods}. Section \ref{sec:deep} delves into state-of-the-art clustering techniques that use deep learning. Section \ref{sec:experiments} provides a comparative experiment of the methods. Finally, Section \ref{sec:applications}  discusses applications of graph clustering.

\section{Background and Definitions} \label{sec:background}

\subsection{Graph Theory}

\subsubsection{Definitions}

\begin{figure}[H]
    \centering
    \includegraphics[width=.5\linewidth]{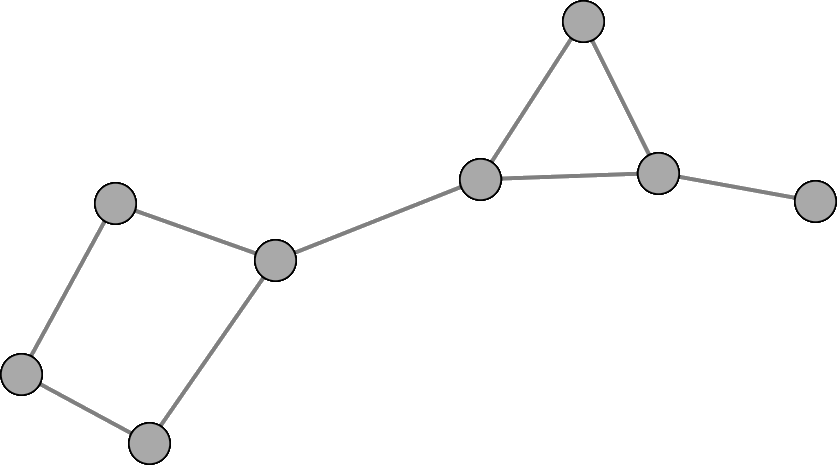}
    \caption{Graph with eight nodes and nine edges}
    \label{fig:graph}
\end{figure}

Graph Theory is the study of graphs, mathematical structures made up of nodes (or vertices) connected by edges that represent relationships between the nodes. Edges can either be undirected, indicating a bidirectional relationship, or directed, indicating a unidirectional relationship from one node to another. Figure \ref{fig:graph} illustrates a small undirected graph.

The underlying network can represent various real-life systems, including road networks, social networks, molecules, and Internet web pages.

A graph can formally be written $G=(V,E)$, with $V$ being a set of nodes and $E$ a set of edges.
In addition, the nodes can contain features, represented in the feature matrix $\mathbf{X}$ of
shape $\lvert V\rvert\times d$ ($d$ being the number of features per node), providing additional
information for each node. Finally, an edge between nodes $i$ and $j$ can be \textit{weighted}
with a real value $w_{ij}$, standing for the weight of the connection.

\subsubsection{Matrix Representation}

Given a graph $G=(V,E)$ we define various matrices that will be used in the latter.

\paragraph{Adjacency Matrix}
The adjacency matrix $\mathbf{A}$ is a fundamental representation of a graph in the field of graph theory. It holds information about the edges by indicating the connections between nodes through binary values, $1$ meaning the presence of an edge, and $0$ the absence of an edge. It is a $n\times n$ diagonal matrix defined as:
\[
A_{i,j}=
    \begin{cases}
        1 & \text{ if } (i,j) \text{ is an edge},\\
        0 & \text{ otherwise}.\\
    \end{cases}
\]

In addition to the symmetric property, the eigenvectors of the adjacency matrix $A$ are real and orthogonal.

\paragraph{Degree Matrix}
The \textit{degree} of a node, denoted as $\text{deg}(i)$ for the node $i$, represents the number of edges incident to the node, i.e., the edges connected to the node. The degree matrix $\mathbf{D}$ is defined as a diagonal $n \times n$ matrix, where $n=|V|$ is the total number of nodes in the graph. Each diagonal element $D_{ii}$ corresponds to the degree of node $i$, and is defined as:

\[
D_{i,j} = 
\begin{cases}
    \text{deg}(i)=\sum_{k=1}^n A_{ik} & \text{if } i = j, \\
    0 & \text{otherwise.}
\end{cases}
\]

\paragraph{Normalized adjacency Matrix}
Furthermore, the normalized adjacency matrix $\mathbb{A}$ is defined as:
\[
\mathbb{A} = \mathbf{D}^{-\sfrac{1}{2}}\mathbf{A} \mathbf{D}^{-\sfrac{1}{2}}
\]
$\mathbf{D}$ being a diagonal matrix, $\mathbf{D}^{-\sfrac{1}{2}}$ can be computed as:
\[
\mathbf{D}^{-\sfrac{1}{2}} =
\begin{pmatrix}
\sfrac{1}{\sqrt{\text{deg}(1)}} & 0 & \cdots & 0 \\
0 & \sfrac{1}{\sqrt{\text{deg}(2)}} & \cdots &0 \\
\vdots & \vdots & \ddots & \vdots \\
0 & 0 & \cdots & \sfrac{1}{\sqrt{\text{deg}(n)}}
\end{pmatrix}
\]

\paragraph{Laplacian Matrix}
For a graph $G=(V,E)$ of adjacency matrix $\mathbf{A}$ and degree matrix $\mathbf{D}$, the Laplacian matrix $\mathbf{L}$ of size $\lvert V\rvert\times\lvert V\rvert$ is defined as: 
\[
    L_{i,j}=
    \begin{cases}
        \text{deg}(i) & \text{ if } i = j ,\\
        -1 & \text{ if } i\neq j \text{ and } A_{ij}=1,\\
        0 & \text{ otherwise.}\\
    \end{cases}
\]
Simply put, $\mathbf{L}=\mathbf{D}-\mathbf{A}$.
This matrix holds interesting mathematical properties such as being positive semi-definite, and it gives indications about the graph's connectivity (e.g. its second eigenvalue measures how well is the graph connected).
 
\subsection{Graph Clustering}

\begin{figure}[H]
    \begin{subfigure}[c]{0.49\linewidth}
        \centering
        \includegraphics[width=\textwidth]{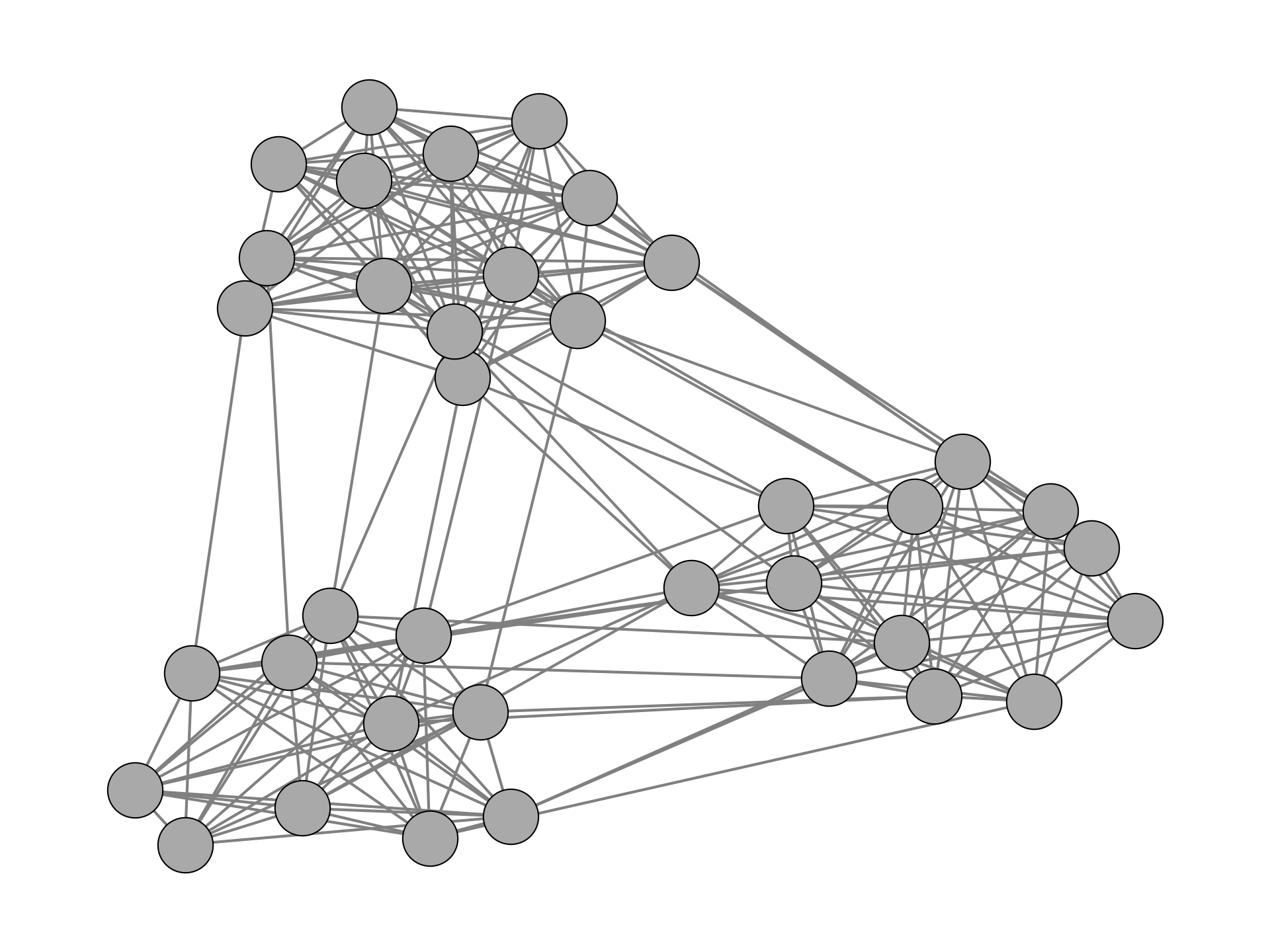}
    \end{subfigure}
    \hfill
    \begin{subfigure}[c]{0.49\linewidth}
        \centering
        \includegraphics[width=\textwidth]{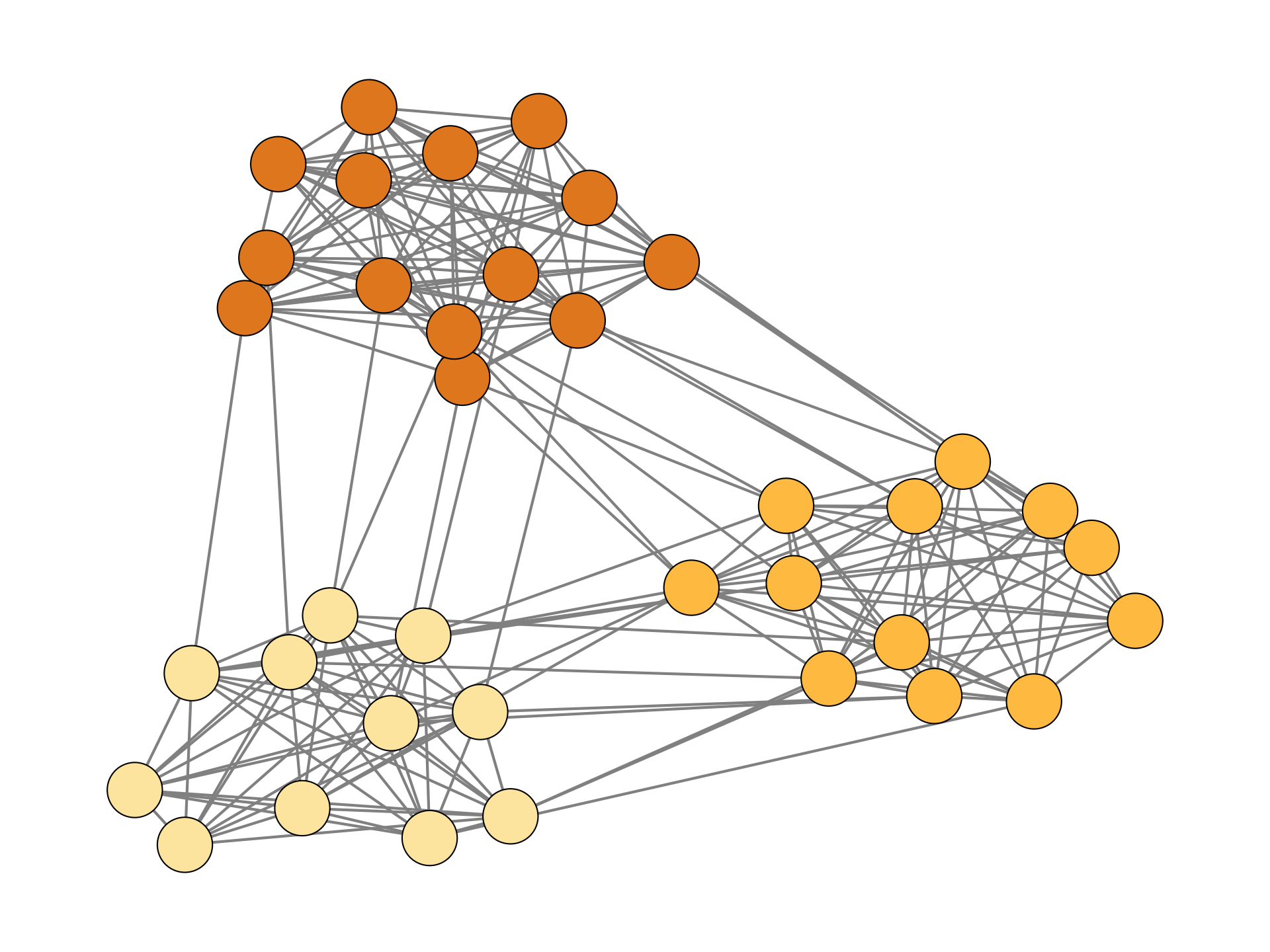}
    \end{subfigure}
    \caption{Graph clustering with three clusters}
    \label{fig:graph_clustering}
\end{figure}

Graph clustering has been extensively researched for several decades and remains a hot topic in contemporary research. This area of study focuses on grouping nodes in a graph into clusters based on various criteria, such as graph structure, node characteristics, or a combination of both. Despite significant progress made, challenges such as scalability (to a high number of nodes), accuracy, and integration of new data types continue to drive ongoing research and development in this field. Figure \ref{fig:graph_clustering} illustrates a clustering process of three communities.

Formally, clustering a graph $G=(V,E)$ is the process of finding an optimal partition $\mathcal{C}=(C_1,C_2,\dots,C_k)$ of the node set $V$, such that:
\begin{itemize}
    \item $C_i\subseteq V$ $\forall i =1,2,\dots,k$ 
    \item $C_i \cap C_j = \emptyset, \quad \space \forall \space i \neq j$
    \item $\bigcup_{i=1}^k C_i = V$
    \item Nodes within each cluster $C_i$ are more densely connected to each other than to nodes in other clusters $C_j$ ($i \neq j$)
\end{itemize}

The optimality is defined through various metrics detailed in the next section.

\subsection{Clustering Evaluation}

Several metrics are used to evaluate a clustering result. The first section presents supervised evaluations, used when provided with the ground truth labels, while the second one presents unsupervised evaluations which can be used without prior knowledge of the real node labels, and for insights beyond known labels.

\subsubsection{Supervised Evaluations}

\paragraph{Accuracy}

The most forward metric is accuracy (ACC). It takes values in $[0,1]$, $1$ as the best, and measures the proportion of correctly classified nodes out of the total number of nodes. It is usually used for supervised learning with the formula:
\[
    \text{ACC}(\mathbf{y},\mathbf{\hat{y}})=\frac{1}{n}\sum^{n-1}_{i=0}1_{y_i=\hat{y}_i} 
\]
where $\mathbf{y}$ are the ground truth labels, $\mathbf{\hat{y}}$ the predicted labels, and $n$ the number of nodes.

However, an issue arises when using the accuracy of unsupervised learning. In fact, there is no association between the real class labels and the clustering output. For example, a ground truth vector of $(1\text{ } 0\text{ } 0)^{T}$ and a clustering result of $(0\text{ } 1\text{ } 1)^{T}$ would give an accuracy of $0$, while being correct. To solve this problem, \cite{morbieu_accuracy_nodate} uses permutations of the clustered labels to find the maximum accuracy achieved with this formula:
\[
    \text{ACC}(\mathbf{y},\mathbf{\hat{y}})=\max_{\text{perm}\in P}\frac{1}{n}\sum^{n-1}_{i=0}1_{\text{perm}(y_i)=\hat{y}_i}
\]
where $P$ is the set of all permutations in $[1,K]$, $K$ being the number of clusters.

\paragraph{Normalized Mutual Information}

Normalized Mutual Information (NMI) takes values in $[0, 1]$ and should be maximized. It is specifically designed for clustering evaluation and measures the similarity between the truth labels and the predicted clusters. NMI evaluates similarity at the level of clusters. It considers how well the groups formed by the clustering algorithm correspond to the true groups in the data. The formula is:
\[
    \text{NMI}=\frac{I(Y,\hat{Y})}{\sqrt{H(Y)\times H(\hat{Y})}}
\]
where $Y$ are the ground truth labels and $\hat{Y}$ the clustering labels. $I$ is the Mutual Information (MI), and $H$ is entropy.

The MI is calculated as:
\[
    I(Y,\hat{Y})=\sum_{y\in Y}\sum_{\hat{y}\in \hat{Y}}p(y,\hat{y})\log \left( \frac{p(y,\hat{y})}{p(y)p(\hat{y})}\right)
\]
where $p(y,\hat{y})$ is the joint probability of $Y$ and $\hat{Y}$, and $p(y)$ and $p(\hat{y})$ are
the marginal probability distributions of $Y$ and $\hat{Y}$, respectively. 

The entropy $H$, which measures the uncertainty of the cluster assignments. It is defined as:
\[
    H(Y)=-\sum_{y\in Y}p(y)\log p(y)
\]

\paragraph{Adjusted Rand Index}

The Adjusted Rand Index (ARI) also measures the similarity between two clustering by considering how much the observed agreement exceeds what would be expected by random chance alone. It ranges from $-1$ to $1$, $1$,
which means that the two clusters are identical. Its calculation starts by computing the contingency table between the predicted partition $\hat{Y}$ and the real partition $Y$. In the table, $n_{ij}$ is the number of elements in common between the partition $i$ and $j$. The contingency table is illustrated in table \ref{tab:contingency}.

Then, the sums of the combinations of $a$, $b$ and $c$ are calculated with:
\begin{align*}
    a = \Sigma_{i} \binom{a_i}{2} &= \Sigma_{i} \frac{a_i (a_i - 1)}{2}\\
    b = \Sigma_{j} \binom{b_j}{2} &= \Sigma_{j} \frac{b_j (b_j - 1)}{2}\\
    c = \Sigma_{i} \Sigma_{j} \binom{n_{ij}}{2} &= \Sigma_{i} \Sigma_{j} \frac{n_{ij}(n_{ij}-1)}{2} 
\end{align*}
where $a$ is the sum of the rows of the table, $b$ is the sum of the columns, and $c$ is the sum of all the combinations.

\begin{table}[H]
    \centering
    \caption{Contingency table between two partitions}
    \begin{tabular}{c|cccc|c}
        \diagbox[dir=SE,height=2em]{\small $Y$}{\small $\hat{Y}$} & $\hat{Y}_1$ & $\hat{Y}_2$ & $\cdots$ & $\hat{Y}_s$ & sums\\
        \hline
        $Y_1$ & $n_{11}$ & $n_{12}$ & $\cdots$ & $n_{1s}$ & $a_1$\\
        $Y_2$ & $n_{21}$ & $n_{22}$ & $\cdots$ & $n_{2s}$ & $a_2$\\
        $\vdots$ & $\vdots$ & $\vdots$ & $\ddots$ & $\vdots$ & $\vdots$\\
        $Y_r$ & $n_{21}$ & $n_{22}$ & $\cdots$ & $n_{rs}$ & $a_r$\\
        \hline
        sums & $b_1$ & $b_2$ & $\cdots$ & $b_s$ & \\
    \end{tabular}
    \label{tab:contingency}
\end{table}

After that, the Expected Index (EI) can be computed, representing the Rand index under the hypothesis that the two clustering come from independent assignments:
\[
    \text{EI} = \frac{a \cdot b}{\binom{n}{2}}
\]

Finally, the ARI is computed through the normalization of the total number of combinations by the EI:
\[
    \text{ARI} = \frac{c - \text{EI}}{\sfrac{1}{2}(a + b) - \text{EI}}
\]

\subsubsection{Unsupervised Evaluations}

\begin{figure}[H]
    \centering
    \includegraphics[width=0.75\linewidth]{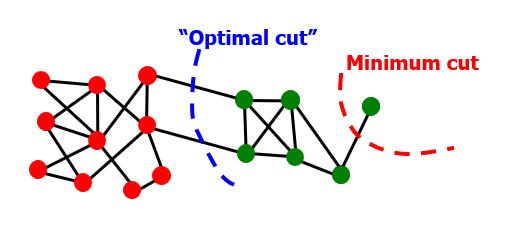}
    \caption{Example of a situation where the minimum cut does not provide the optimal solution \cite{leskovec2019}}
    \label{fig:mincutsuboptimal}
\end{figure}

\paragraph{Cut Score} The Cut Score is a metric used when ground truth labels are not provided. Graph clustering can be performed by minimizing the cut score, which is defined for a cluster $A$ as:
\[
    \text{Cut}(A) = \sum_{i \in A, j \notin A} w_{ij}
\]
where $w_{ij}$ is the weight of the edge between nodes $i$ and $j$. If $G$ is unweighted, $w_{ij} = 1$.  In this case, the quality of a cluster is the weight of connections pointing outside the cluster.\\
However, minimizing the cut score does not always produce the best solution, as illustrated in Figure \ref{fig:mincutsuboptimal}.

\vspace*{-1\baselineskip}

\begin{figure}[H]
    \centering
    \includegraphics[width=0.65\linewidth]{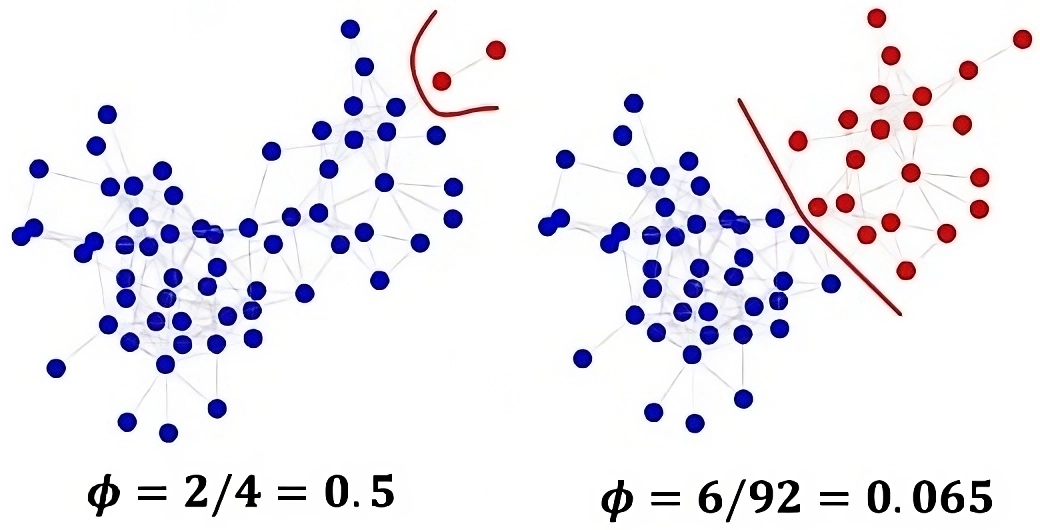}
    \caption{Comparison between cut score and conductance metrics \cite{leskovec2019}}
    \label{fig:cutvsconduct}
\end{figure}

\paragraph{Conductance} To address the limitations of the cut score, a better unsupervised metric is used, called conductance. Conductance measures the connectivity of the cluster to the rest of the graph relative to the density of the cluster itself:
\[
    \phi(A) = \frac{\{(i,j) \in E ; i \in A, j \notin A\}}{\min(\text{vol}(A))} = \frac{\text{Cut}(A)}{\min(\text{vol}(A))} 
\]
where $\text{vol}(A)$ is the total weight of the edges with at least one endpoint in $A$, defined as:
\[ 
    \text{vol}(A) = \sum_{i \in A} d_i
\]
where $d_i$ the degree of node $i$. Due to the density parameter, the resulting clusters are more balanced compared to using the cut score, as shown in Figure \ref{fig:cutvsconduct}.

\vspace*{-1\baselineskip}
\paragraph{Internal Evaluation Measures}  \label{par:modularity}

Internal evaluation measures assess the quality of clusters based on the information intrinsic to the graph. One widely used internal measure is the \textit{modularity} \(Q\), which is defined as:
\begin{equation} \label{eq:modularity}
    Q = \frac{1}{2m} \sum_{i,j} \left[ A_{ij} - \frac{k_i k_j}{2m} \right] \delta(c_i, c_j)
\end{equation}
where \(\mathbf{A}\) is the adjacency matrix of the graph, \(k_i\) and \(k_j\) are the degrees of vertices \(i\) and \(j\), \(m\) is the total number of edges, \(c_i\) and \(c_j\) are the clusters of vertices \(i\) and \(j\), and \(\delta(c_i, c_j)\) is 1 if \(i\) and \(j\) are in the same cluster, and 0 otherwise. High modularity indicates strong intra-cluster connections and weak inter-cluster connections.

\paragraph{Internal Density}

Internal density measures the cohesiveness of the clusters within a graph. It measures the ratio between the number of edges in a cluster $C$ and the maximum number of possible edges within $C$. For a single cluster, it is defined as
\[
    \rho (C)=\frac{m_C}{\frac{n_C(n_C-1)}{2}}
\]
where $m_c$ is the number of edges within the cluster $C$ and $n_c$ the number of nodes in $C$.
Values close to one indicate that the cluster is densely connected, while low values close to zero indicate poor connections.

For the complete graph, the measure uses a weighted average of all internal densities:
\[
    \text{Internal Density}=\frac{\sum_{i=1}^kn_{C_i}\times \rho(C_i)}{\sum_{i=1}^kn_{C_i}}
\]
where $C_i$ is the cluster $i$, $n_{C_i}$ the number of nodes in the cluster $i$.

\subsection{Deep Learning}

Deep Learning methods for graph clustering have been extensively used in the last ten years, due to their powerful nonlinear representation capacity. They are mostly built on top of Autoencoders and Graph Neural Networks. For this reason, it is necessary to understand these concepts before diving into Deep Graph Clustering. 

\subsubsection{Artificial Neural Networks}

\begin{figure}[H]
    \centering
    \includegraphics[width=.5\linewidth]{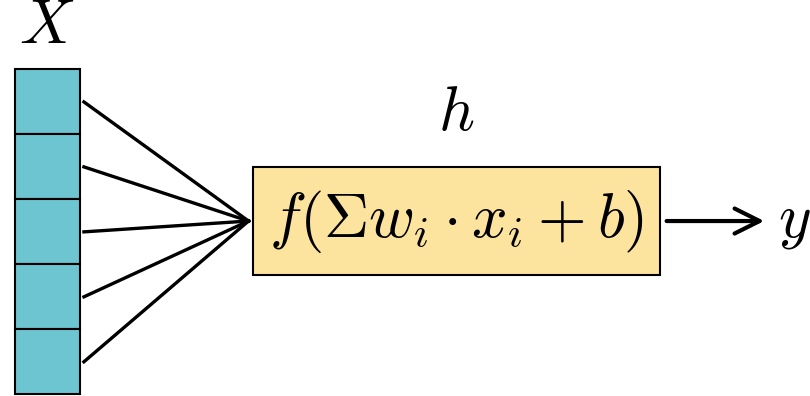}
    \caption{Single neuron operation}
    \label{fig:neuron}
\end{figure}

An Artificial Neural Network (ANN), also called a Multilayer Perceptron (MLP), is a composition of neurons organized in several connected layers. Each neuron receives information and applies a linear function to the input, a weighted linear combination with a bias added. An activation function (e.g. $\text{tanh}$, $\text{ReLU}$) is then applied to the output for nonlinearity (illustrated in Figure \ref{fig:neuron}).

\begin{figure}[H]
    \centering
    \includegraphics[width=.45\linewidth]{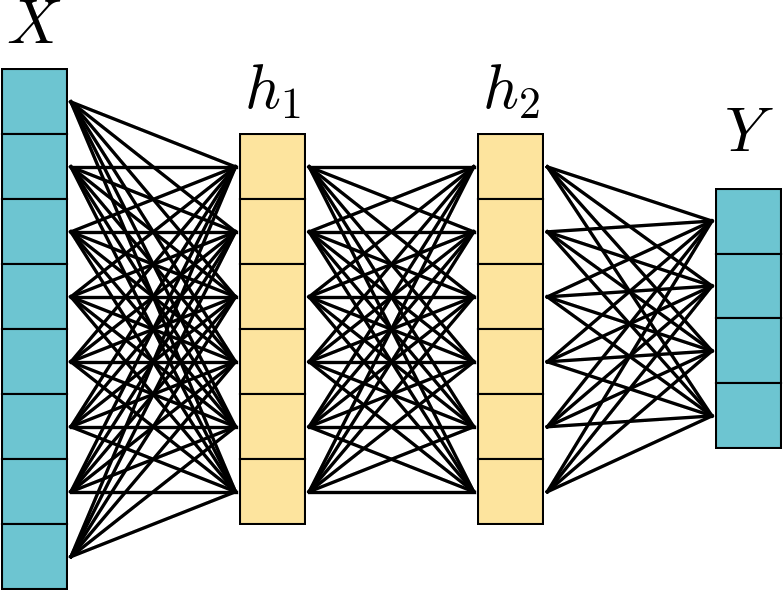}
    \caption{Neural Network with two hidden layers}
    \label{fig:neuralnet}
\end{figure}

Each layer of the network is composed of several neurons (illustrated in Figure \ref{fig:neuralnet}), enabling the network to learn complex non-linear functions. Each layer can be written as a matrix operation. For a layer with input size $n$ and output $m$, the layers are composed of a $n\times m$ weight matrix $\mathbf{W}$, a bias $\mathbf{B}$ of size $1\times m$, and an activation function $f$. The output $\mathbf{Y}$ of the layer is calculated as:
\[
    \mathbf{Y}=f(\mathbf{WX} + \mathbf{B})
\]

The weights of the neurons are learned through a process called training. For each input $X$, there is an associated desired output $Y$. To update the weights of the network, a loss function $\mathcal{L}(\mathbf{Y}, \mathbf{\hat{Y}})$, which quantifies the difference between the calculated result and the desired result, is used in conjunction with a gradient descent algorithm for back-propagation through the entire network. This process is executed iteratively until the neural network converges.

\subsubsection{Autoencoders} \label{sssec:autoencoder}

\begin{figure}[H]
    \centering
    \includegraphics[width=.5\linewidth]{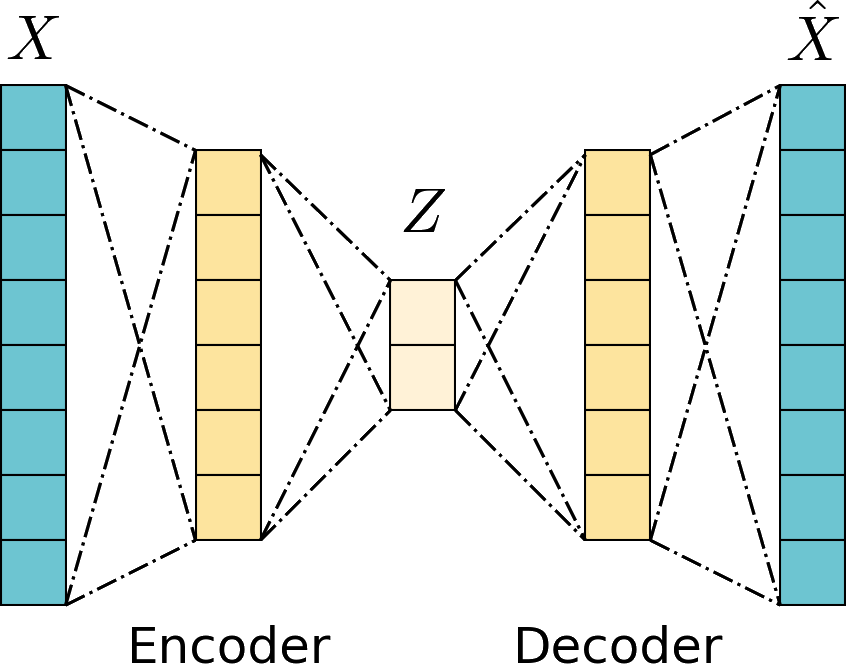}
    \caption{The input vector $\mathbf{X}$ is encoded into the latent vector $\mathbf{Z}$, and decoded back to $\hat{\mathbf{X}}$}
    \label{fig:autoencoder}
\end{figure}

Autoencoders (AE), introduced in 2006 by \cite{Hinton2006}, are a particular neural network architecture in unsupervised learning that is primarily used for data compression. It uses a neural network encoder to reduce the dimensionality of input data $\mathbf{X}$, to obtain the latent vector representation $\mathbf{Z}$, and a symmetric architecture from the decoder to reconstruct the original vector $\hat{\mathbf{X}}$ (illustrated in Figure \ref{fig:autoencoder}). The encoder and decoder architectures can vary widely, spanning from shallow fully connected layers to deep convolutional neural network layers, and can be applied to various data types such as vectors, time series, and images.

One of the most important advantages of this model compared to standard dimensionality reduction methods -- for instance Principal Component Analysis (PCA) -- is the ability of the Neural Networks to learn non-linear relationships.

The loss of the autoencoder generally uses the Mean Square Error (MSE):
\[
    \mathcal{L}_{\text{AE}}(\mathbf{X},\hat{\mathbf{X}}) = \frac{1}{n}\sum_{i=0}^{n}(\mathbf{x_i} - \mathbf{\hat{x}_i})^2
\]
where $\mathbf{X}$ are the input data and $\hat{\mathbf{X}}$ the reconstructed data. This architecture has achieved outstanding performance and precision, and has also been used for data denoising \cite{autoencoder_denoising}, anomaly detection \cite{autoencoder_anomaly} and even for data generation \cite{autoencoder_gen}.

\subsubsection{Graph Convolutional Networks}

\begin{figure*}[!tp] 
    \centering
    \includegraphics[width=.85\linewidth]{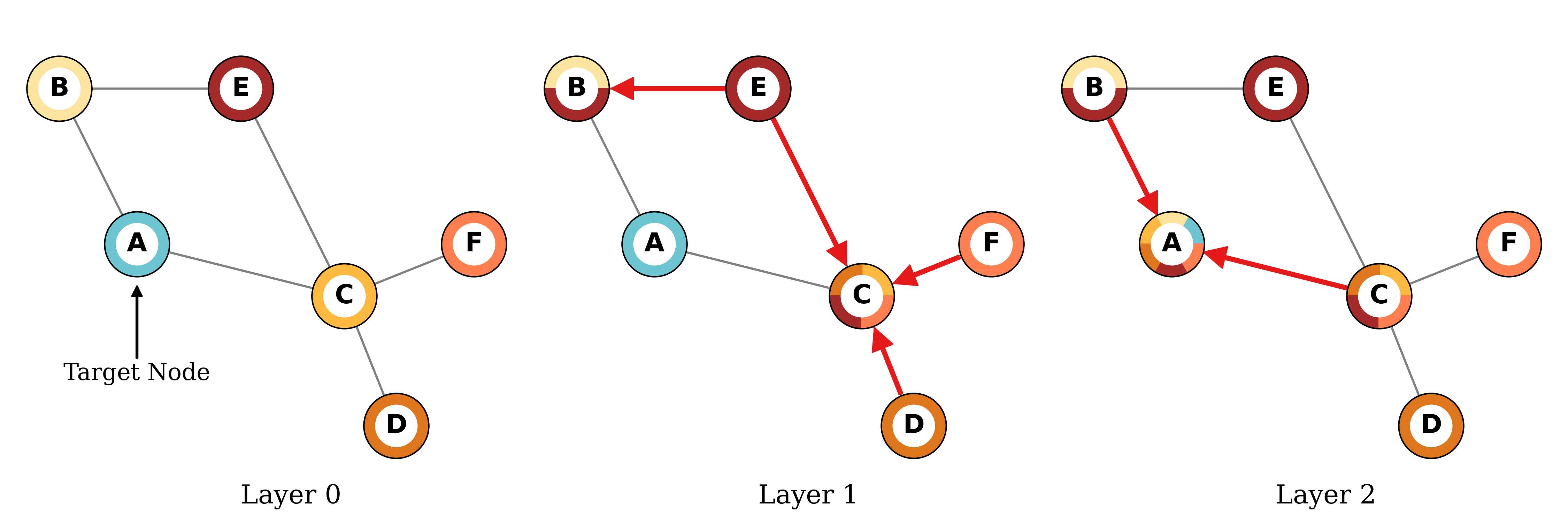}
    \caption{Simplified illustration of a two-layer message passing to target node $A$ (messages are represented by red arrows). The first layer passes information from second-degree neighbors of $A$ to the neighbors of $A$, and in the second layer, $A$ receives information from its neighbors, thereby it receives information from the whole graph.}
    \label{fig:messagepassing}
\end{figure*}

In recent years, Graph Neural Networks (GNNs) have emerged as a powerful tool for processing and analyzing
natural-representative data as graphs \cite{Zhang_Cui_Zhu_2020,Wu_Pan_Chen_Long_Zhang_Yu_2021,Zhou_Cui_Hu_Zhang_Yang_Liu_Wang_Li_Sun_2020}. Unlike traditional deep learning methods that excel with grid-like data structures such as images, text, audio, and video, GNNs are designed to handle the complex relationships and interactions found in graph data. They are part of what is called \textit{Geometric Deep Learning}, which studies data beyond Euclidean space, such as grids, groups, and geodesics \cite{Bronstein_Bruna_Cohen_Veličković_2021, Bronstein_Bruna_LeCun_Szlam_Vandergheynst_2017}.

Inspired by Convolutional Neural Networks (CNNs), which operate on one- and two-dimensional data by capturing the context of each data point with its surroundings (e.g. surrounding pixels in an image), graph Convolutional Networks (GCNs) aim to capture the context of a node within its neighborhood. GCNs are organized into layers, where each node is updated based on the connected nodes.

\paragraph{Message Passing} 
The fundamental components of Graph Convolutional Networks (GCNs) are message passing operators \cite{Bronstein_Bruna_Cohen_Veličković_2021}. Consider a graph $G=(V,E)$ with node features $\mathbf{X}$. The feature vector $\mathbf{x_u}$ of node $u$ is updated to $\mathbf{h_u}$ in the next layer through a message passing operator, defined as:
\[
    \mathbf{h_u}=\phi\left(\mathbf{x_u},\bigoplus_{v\in \mathcal{N}_u}\psi\left(\mathbf{x_u},\mathbf{x_v}, \mathbf{e_{uv}}\right)\right)
\]
where $\mathcal{N}_u$ is the neighborhood of $u$ (the nodes connected to $u$ via an edge), $\psi$ and $\phi$ differentiable functions, (e.g. neural networks), $\mathbf{e_{uv}}$ the features, if available, of the edge $(u,v)\in E$, and $\bigoplus$ a permutation-invariant aggregation operator such as sum, mean or max. At each iteration, a node receives information about its neighbors. With one layer, the node receives information about its neighbor. A second layer enables the node to receive information from the neighbors of its neighbors. If repeated for $n$ layers, the node receives information from its neighbors at the $n$-th degree. Figure \ref{fig:messagepassing} illustrates a simplified visualization of a two-layer message passing to a target node.

\paragraph{Convolutional Operator}
The convolutional operator was introduced in 2016 by \cite{Kipf_Welling_2017}. It uses message passing with a trainable weight matrix, where the updated node features $\mathbf{h_u}$ are defined, for unweighted graphs, as:
\[
    \mathbf{h_u}=\sigma \left(\mathbf{\Theta} ^\top \sum_{v\in (u\cup \mathcal{N}_u)} \frac{1}{\sqrt{\tilde{d_u}\tilde{d_v}}}\mathbf{x_v}\right)
\]
with $\mathbf{\Theta}$ the trainable weight matrix, $\sigma$ an activation function, and $\tilde{d_u}=\text{deg}(u) + 1$. It is important to note that this operator adds self-loops to each node to receive information from the node itself. 

This operator can also be written as a matrix operation for the complete graph:
\[
    \mathbf{H} = \sigma\left( \mathbf{\tilde{D}}^{-\sfrac{1}{2}}\mathbf{\tilde{A}}\mathbf{\tilde{D}}^{-\sfrac{1}{2}}\mathbf{X}\mathbf{\Theta} \right)
\]
where $\mathbf{H}$ is the updated feature matrix, $\sigma$ an activation function.
$\mathbf{\tilde{D}}$ and $\mathbf{\tilde{A}}$ are, respectively, the degree matrix and adjacency matrix with added self-loops ($\mathbf{\tilde{D}}=\mathbf{D}+\mathbf{\text{diag}}(1,1,\dots,1)$, $\mathbf{\tilde{A}}=\mathbf{A}+\mathbf{\text{diag}}(1,1,\dots,1)$).

GCNs have achieved high performance in various tasks, including nodes classification \cite{Kipf_Welling_2017, Yao_Mao_Luo} and time series forecasting \cite{Zhao_Song_Zhang_Liu_Wang_Lin_Deng_Li_2020, Bai_Zhu_Song_Zhao_Hou_Du_Li_2021}.

\section{Traditional Clustering} \label{sec:tradmethods}

Traditional graph clustering techniques have laid the foundation for understanding complex network structures by grouping nodes into meaningful clusters or communities based on the graph structure. In this section, several prominent traditional graph clustering methods are explored, and their theoretical foundations are detailed. The most popular algorithm, Spectral Clustering, is presented first, introducing the notion of the spectrum and the associated clustering method. Next, Stochastic Block Models are detailed, introducing a probabilistic approach and focusing on the differences between various versions of the associated clustering algorithms. Following this, the Markov Clustering Algorithm is explained, utilizing random walks and Markov chains to iteratively reveal clusters within a graph. Lastly, the Leiden Algorithm is introduced, recognized for its superior performance in detecting community structures in large-scale networks.

\subsection{Spectral Clustering} \label{ssec:spectral}

Spectral clustering \cite{von_Luxburg_2007} is a powerful method for partitioning graph data into clusters based on the eigenvalues and eigenvectors of a similarity matrix. Unlike standard clustering methods, spectral clustering can uncover complex structures in data with the graph structure, making it valuable for various machine learning and data analysis tasks. In this part, the theory, applications, and advantages of spectral clustering are explored, providing insights into its mathematical underpinnings and practical implications.

\subsubsection{Important Properties of the Normalized Laplacian Matrix}

The Laplacian matrix is always positive semi-definite, the smallest eigenvalue of this matrix is always 0 since its rows sum up to zero. \\
The multiplicity of the zero eigenvalues is equal to the number of connected components of the graph. 

Furthermore, the normalized Laplacian matrix $\mathbb{L}$ is defined as: 
\begin{align*}
    \mathbb{L} &= \mathbf{I} - \mathbb{A} = \mathbf{D}^{-\sfrac{1}{2}}(\mathbf{D}-\mathbf{A}) \mathbf{D}^{-\sfrac{1}{2}}\\
    \mathbb{L} &= \mathbf{D}^{-\sfrac{1}{2}} \mathbf{L} \mathbf{D}^{-\sfrac{1}{2}}
\end{align*}
If $\mathbb{L}$ is used, all eigenvalues are less than or equal to 2, and the largest eigenvalue equals 2 if and only if one of the connected components of the graph is bipartite.

The number of connected components is the maximum number of disjoint subsets of $G$'s vertices, such that there are no edges connecting vertices of distinct subsets. In terms of the adjacency matrix $\mathbf{A}$, the number of connected components of $G$ is the maximum number of diagonal blocks that can be achieved by the same permutation of the rows and columns of $\mathbf{A}$.

It is important to note that if $G$ is connected, then 0 is a single eigenvalue with a corresponding unit-norm eigenvector:
\[
    \mathbf{u_{0}} = \frac{1}{\sqrt{n}} 
    \begin{pmatrix}
      1 \\
      1 \\
      \vdots \\
      1 \\
    \end{pmatrix}
\]

\subsubsection{Notion of Spectrum}
The spectrum of a matrix is defined as the set of its eigenvalues $\lambda_i$, each associated with a corresponding eigenvector $\mathbf{u_i}$. In spectral clustering, this concept is used by decomposing a matrix, often the Laplacian of a graph, into its spectral components. This decomposition provides a new representation of the nodes, which can be used to identify clusters by applying standard clustering techniques.

\subsubsection{Finding the Eigenvalues}
The search for the second eigenvalue $\lambda_2$ can be formulated as an optimization problem using the
Min-Max Courant-Fisher theorem \cite{courant1989}. The second-smallest eigenvalue $\lambda_2$ is given by:
\begin{equation} \label{eq:courantfischer}
\lambda_2 = \min_\mathbf{x} \frac{\mathbf{x}^\top \mathbf{L} \mathbf{x}}{\mathbf{x}^\top \mathbf{x}}
\end{equation}
where $\mathbf{L}$ is the Laplacian matrix and $\mathbf{x}$ a vector. Each element $x_i$ of $\mathbf{x}$ represents the value assigned to node $i$ in this new representation of the graph.

Calculating $\mathbf{x}^\top L\mathbf{x}$ reveals that:
\[
    \mathbf{x}^\top \mathbf{L} \mathbf{x} = \sum_{(i,j) \in E} (x_i - x_j)^2
\]
This relationship, known as the Rayleigh Theorem, is crucial because it provides a way to express the optimization problem in terms of the graph's edges.

\begin{figure}[H]
    \centering
    \includegraphics[width=0.6\linewidth]{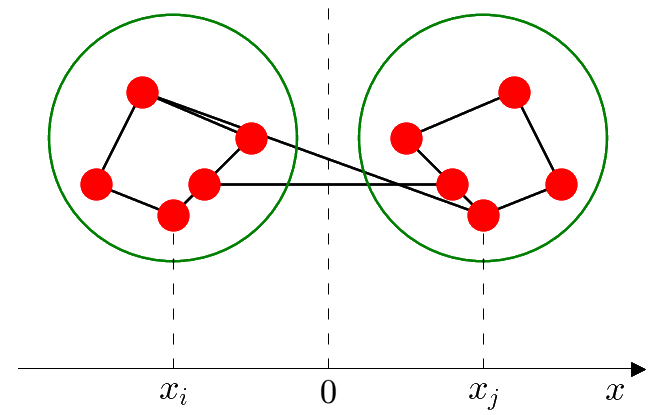}
    \caption{Minimization of $x^\top L x$ }
    \label{fig:min_pb}
\end{figure}

To minimize $\mathbf{x}^\top L \mathbf{x}$, the values $x_i$ are assigned to the nodes such that the sum of squared differences $(x_i-x_j)^2$ for edge $(i,j)$ is minimized. The goal is to find a configuration in which few edges cross the zero value of $x$. This is based on the constraint derived from eq.  \eqref{eq:courantfischer} that the sum of the node labels should be zero, intuitively resulting in two clusters of roughly equal size.

Thus, in order to minimize:
\begin{equation} \label{eq:sumsq}
\sum_{(i, j) \in E} (x_i - x_j)^2
\end{equation}
the number of edges between different clusters should be minimized, while edges within the same cluster will naturally contribute less to the sum. Figure \ref{fig:min_pb} illustrates this optimization problem.

\subsubsection{Spectral Clustering Algorithm}

The complete method, as described in \cite{Shi_Malik_2000}, is presented in Algorithm \ref{alg:spectral}.

\begin{algorithm}[H]
\caption{Spectral Clustering Algorithm using multiple eigenvectors}
\label{alg:spectral}
\SetAlgoLined
\KwIn{Graph $G = (V, E)$ with adjacency matrix $\mathbf{A}$ and degree matrix $\mathbf{D}$, number of clusters $k$}
\KwOut{Clusters assignments $\mathcal{C} = (C_1,C_2,\dots, C_k)$}
\BlankLine

Compute the normalized Laplacian matrix $\mathbb{L}$\\
    Compute the first $k$ eigenvectors $\mathbf{u_k}$ and eigenvalues $\lambda_k$ of $\mathbb{L}$ thanks to the Courant-Fisher theorem.\\
    Construct the matrix $\mathbf{U} \in \mathbb{R}^{n\times k}$ with the vectors $\mathbf{u_k}$ as columns.\\
    Let $\mathbf{y_i} \in \mathbb{R}^{k}$ be the vector corresponding to the $i$-th row of $\mathbf{U}$, representing node $i$. \\
    For each $\mathbf{y_i}$ group the points $(\mathbf{\mathbf{y_i}})_{i=1,\dots ,n}$ with a classical clustering algorithm into clusters $\mathcal{C} = (C_1,C_2,\dots, C_k)$
\end{algorithm}

\paragraph{Finding the Optimal Clusters Number}
Determining the optimal clusters number $k$ can be quite a challenging task.  Two main approaches are possible.

The first method uses recursive bipartitioning \cite{Hagen_Kahng_1992}. This method recursively applies a bi-partitioning algorithm in a hierarchical divisive manner by just using $\lambda_2$ (the second-smallest eigenvalue of the normalized Laplacian matrix). However, this method is inefficient and unstable.

The second method uses multiple eigenvectors \cite{Shi_Malik_2000}. A reduced space is built from the first $n$ eigenvectors and a standard algorithm such as $k$-means is applied. It is a preferable and more efficient solution.

\subsubsection{Spectral Clustering Illustrations}

\begin{figure}[H]
    \centering
    \includegraphics[width=0.9\linewidth]{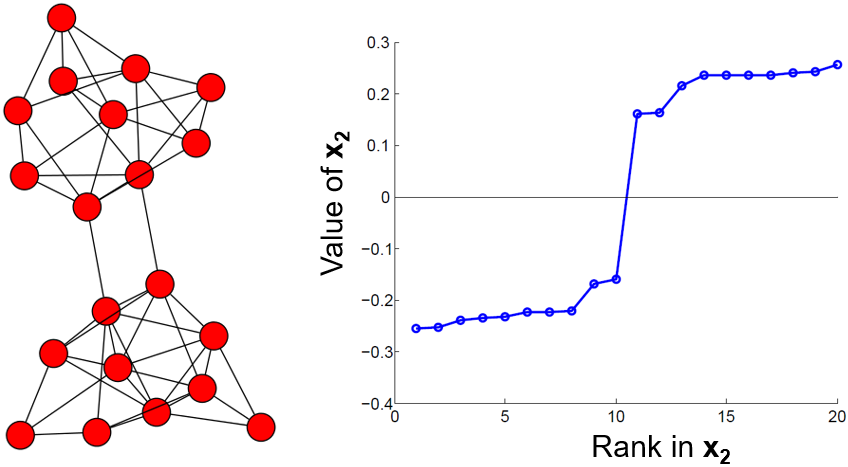}
    \caption{Example of Spectral clustering on a graph with two clusters \cite{leskovec2019}}
    \label{fig:spectral2}
\end{figure}

In Figure \ref{fig:spectral2}, the gap at 0 can be observed and a standard clustering algorithm such as $k$-means can be applied.

\begin{figure}[H]
    \centering
    \includegraphics[width=0.9\linewidth]{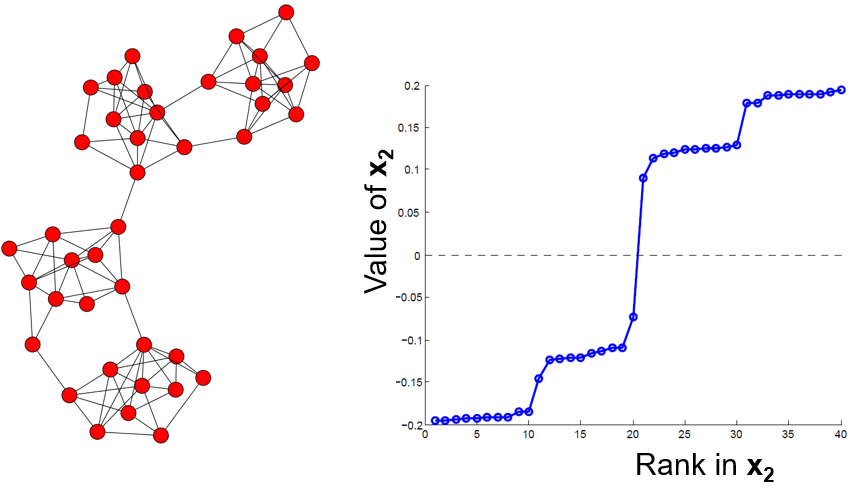}
    \caption{Example of Spectral clustering on a graph with four clusters \cite{leskovec2019}}
    \label{fig:spectral4}
\end{figure}

Similar gaps at three values of $x_2$ can be observed in Figure \ref{fig:spectral4}, giving four distinct clusters.

\subsection{Stochastic Block Models}

Stochastic Block Models (SBM) \cite{Holland_Laskey_Leinhardt_1983,Lee_Wilkinson_2019} constitute a category of probabilistic models designed to analyze complex graphs that possess underlying (or latent) structures. In these models, the nodes in the graph are divided into blocks (or communities), with the density of edges between each block determined by a probability distribution. These communities form clusters, and the nodes within each cluster share similar characteristics.

\subsubsection{Stochastic Equivalence}

SBMs exist in various types, each tailored to capture different aspects of graph structures. A fundamental concept in SBM is \textit{stochastic equivalence}, which refers to the idea that nodes within the same block or community have statistically equivalent connection patterns to nodes in other blocks. This equivalence does not imply that the nodes have identical connections but rather that the probability of forming a connection between any two nodes depends only on their block memberships.

\begin{figure}[H]
    \centering
    \includegraphics[width=\linewidth]{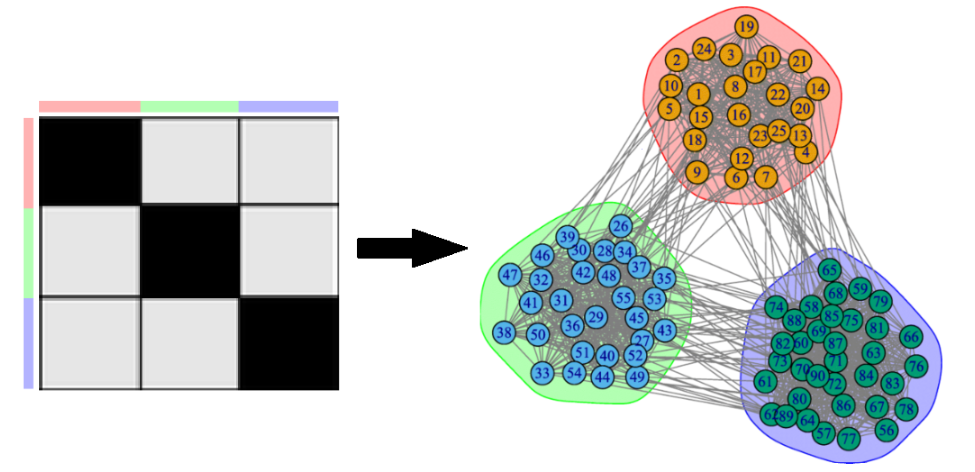}
    \caption{SBM with Block Matrix \cite{Lee_Wilkinson_2019}}
    \label{fig:sbm&matrix}
\end{figure}

A graph that adheres to the concept of stochastic equivalence can be represented by a block matrix $\mathbf{B}$ (Figure \ref{fig:sbm&matrix}), with dimensions $K \times K$ where $K$ represents the number of communities. This matrix allows for the visualization and quantification of interactions between different communities within the graph. Each element $B_{ij}$ in this matrix indicates the probability that a node in block $i$ will connect to a node from block $j$. This facilitates the analysis of connection patterns and the understanding of latent structures within the graph. This matrix representation is essential for modeling the complex relationships between different communities and for predicting future interactions in the graph.

\subsubsection{Standard SBM}

A standard SBM is a statistical model that relies on the notion of stochastic equivalence to analyze graph structures. When inference is conducted on this model, the objective is to estimate essential parameters, such as the probabilities of connection between different blocks and the distribution of nodes within these blocks. This can be accomplished using methods such as maximum likelihood estimation or Bayesian inference. By estimating the parameters of the distribution, the membership to a cluster of each node can be identified, as well as the relationship between each cluster.

\subsubsection{Maximum Likelihood Estimation}
Maximum Likelihood Estimation (MLE) is used to estimate the parameters of a statistical model. In the context of SBM, MLE aims to find the parameters that maximize the likelihood of the observed graph data given the model. This process involves two primary steps.

First, the likelihood function for the SBM is defined. This function represents the probability of observing the given graph data as a function of the model parameters.

Then, optimization techniques, presented in the following sections, are used to find the distribution parameters that maximize this likelihood function.

\paragraph{Likelihood Function}
For a given graph \(G = (V, E)\) with \(N\) nodes with an adjacency matrix \(\mathbf{A}\), suppose that the nodes are divided into \(K\) communities (or clusters). The block matrix \(\mathbf{B}\) of size \(K \times K\) contains the probabilities of connections between nodes in different communities and \(\mathbf{z}\) is the community membership vector where \(z_i\) indicates the community of node \(i\).

The likelihood function \(L(\theta)\) for SBM, where \(\theta\) represents the parameters \((\mathbf{B}, \mathbf{z})\), can be written as:
\begin{align}  \label{eq:sbmlh}
    \begin{split}
    L(\theta) &= P(\mathbf{A} \mid \mathbf{B}, \mathbf{z}) = \prod_{i < j} P(A_{ij} \mid \mathbf{B}, \mathbf{z})\\
    L(\theta) &= \prod_{i < j} B_{z_i z_j}^{A_{ij}} (1 - B_{z_i z_j})^{1 - A_{ij}}
    \end{split}
\end{align}
as each term \(P(A_{ij} \mid \mathbf{B}, \mathbf{z})\) is a Bernoulli probability defined by the elements $B_{ij}$ of the block matrix $\mathbf{B}$.

In concrete terms, the total likelihood of the graph is the product of the probabilities for each possible pair of nodes. This product reflects the joint probability that the SBM model generates all the observed connections and non-connections exactly as in $\mathbf{A}$.

\paragraph{Optimization}

Maximizing this likelihood function directly can be challenging due to the combinatorial nature of community assignments. Therefore, iterative algorithms such as the Expectation Maximization (EM) algorithm are often used to find the MLE.

\subsubsection{Expectation-Maximization Algorithm}
The EM algorithm is an iterative technique employed to maximize the likelihood estimates for the parameters of statistical models, especially when these models involve unobserved latent variables, such as community memberships. The algorithm consists of four steps:


\paragraph{Initialization} The algorithm starts with initial guesses for the parameters \(\mathbf{B}\) and \(\mathbf{z}\), which can be random assignments.

\paragraph{Expectation} In the expectation step (\textit{E}), the expected value of the likelihood function is calculated with respect to the current estimate of the distribution of the latent variables $\mathbf{B}$ and $\mathbf{z}$. This involves estimating the probability that each node belongs to each community based on the current parameters.
    
For each node \(i\), the probability \(\gamma_{ik}\) that it belongs to the community \(k\) can be computed as:
    
    \[
    \gamma_{ik} = P(z_i = k \mid \mathbf{A}, \mathbf{B})
    \]

\paragraph{Maximization} In the maximization step (\textit{M}), the parameters \(\mathbf{B}\) and \(\mathbf{z}\) are updated to maximize the expected likelihood found in the E-step. This involves updating the block matrix \(\mathbf{B}\) and the community membership vector \(\mathbf{z}\).
    
The block matrix \(\mathbf{B}\) is updated based on the current probabilities of community membership, and each probability of connection between two communities $k$ and $l$ is computed as:
    
\[
    B_{kl} = \frac{\sum_{i \neq j} \gamma_{ik} \gamma_{jl} A_{ij}}{\sum_{i \neq j} \gamma_{ik} \gamma_{jl}}
    \]
where $\gamma_{ik}$ ($\gamma_{jl}$) is the probability that the node $i$ belongs to $k$ (respectively node $j$ to community $l$). $\mathbf{A}$ is the adjacency matrix and $\sum_{i \neq j}$ the summation over all distinct pairs of nodes.
    
\paragraph{Convergence} The \textit{E} and \textit{M} steps are repeated until the parameters converge to stable values.

This alternating approach between \textit{E} and \textit{M} transforms a difficult combinatorial problem into a series of more manageable subproblems, allowing efficient convergence to an optimal solution.

\subsubsection{Bayesian Inference}
Bayesian inference is another powerful method for estimating the parameters of SBMs. This approach, based on Bayes' theorem, involves updating prior beliefs with observed data to obtain a posterior distribution of the model parameters.

\paragraph{Bayes Theorem}
The Bayes Theorem states that:
\begin{equation} \label{eq:bayestheorem}
P(\theta|\mathbf{A}) = \frac{P(\theta)}{P(\mathbf{A})} P(\mathbf{A}|\theta)
\end{equation}
where $\theta$ are the model parameters and $\mathbf{A}$ the adjacency matrix. $P(\theta|\mathbf{A})$ corresponds to the posterior probability, which quantifies the probability of the parameters $\theta$ given the observed adjacency matrix $A$. $P(\theta)$ is the prior probability, representing our assumptions or existing knowledge about the parameters before observing the data. $P(\mathbf{A}|\theta)$ is the likelihood (see Eq. \eqref{eq:sbmlh}). Finally, $P(\mathbf{A})$ is the marginal probability of the adjacency matrix, which normalizes the probability.

\paragraph{Definition of Priors}
The first step is to define the prior distributions of the key model parameters, namely the partition of the nodes and the block matrix. There are different ways to define the priors.\newline
A commonly used prior for the partition is the Dirichlet distribution, $\pi \sim Dirichlet(\alpha)$, where $\alpha$ is a vector of concentration parameters, indicating our initial beliefs about the relative size of the blocks.\newline
For the block matrix $\mathbf{B}$, a beta prior is often used for the elements of this matrix, as it allows modeling the link probabilities flexibly. For example, each element $B_{kl}$ can be defined as $B_{kl} \sim Beta(\beta_{1}, \beta_{2})$, where $\beta_{1}$ and $\beta_{2}$ are the parameters of the beta distribution, reflecting our initial beliefs about the density of links between blocks $k$ and $l$.\newline
By defining these priors, prior information and assumptions are incorporated into the model, which helps guide Bayesian inference and manage uncertainty systematically.

\paragraph{Calculating the Posterior Distribution}
The posterior distribution is derived by combining the priors and the likelihood of the data using Eq. \eqref{eq:bayestheorem}. However, calculating the marginal probability $P(\mathbf{A})$ often presents a significant challenge. This quantity is obtained by integrating the likelihood multiplied by the prior:
\[
    P(\mathbf{A}) = \int P(\mathbf{A} \mid \theta) P(\theta) \, d\theta
\]

In complex models such as SBM, this integral becomes intractable due to the high dimensionality and complexity of the distributions involved. In practice, techniques such as Markov Chain Monte Carlo are used to estimate this value.

\subsubsection{Markov Chain Monte Carlo}
Markov Chain Monte Carlo (MCMC) methods provide an effective way to overcome these difficulties by offering a means to sample from the posterior distribution. Concretely, MCMC generates a Markov chain whose stationary state corresponds to the posterior distribution of the parameters. By generating successive samples, MCMC algorithms, such as Metropolis-Hastings, enable the construction of an empirical representation of this distribution. This approach not only allows for the calculation of point estimates of the parameters but also quantifies uncertainty by constructing credible intervals and performing comprehensive Bayesian inference.


The Metropolis-Hastings algorithm is an MCMC technique that allows sampling from complex probability distributions without the need to know their normalization constant (in this case, $P(\mathbf{A})$). It works by proposing changes to the current state of model parameters, which are either accepted or rejected based on a criterion that evaluates how well the new sample adheres to the desired distribution:

\begin{equation}
\label{eq:rationposteriorprob}
\frac{p(\theta' \mid \mathbf{A})}{p(\theta \mid \mathbf{A})} = \frac{p(\mathbf{A} \mid \theta')p(\theta')}{p(\mathbf{A} \mid \theta)p(\theta)}
\end{equation}

Each step depends on the previous one, forming a Markov chain that converges over time to the posterior distribution, thereby facilitating statistical inference in situations where direct calculation of the marginal probability is otherwise too complex or impossible. The output of this algorithm is a sequence of samples drawn from the target distribution. These samples, accumulated over iterations, reflect the distribution and can be used to estimate statistics or make inferences about the analyzed data. The algorithm is presented in Algorithm \ref{alg:metropolis}.

\begin{algorithm}[H]
\caption{Metropolis-Hastings Algorithm}
\label{alg:metropolis}
\SetKwInput{KwIn}{Input}
\SetKwInput{KwOut}{Output}
\SetKwInput{KwData}{Data}
\SetKwInput{KwResult}{Result}

\KwIn{Target distribution $p(x)$, proposal distribution $q(x' \mid x)$, initial state $x_0$, number of iterations $e$.}
\KwOut{Sequence of samples from $p(x)$.}
\BlankLine
Initialize the state $x = x_0$\\
\For{$i = 1 \dots e$}{
    Propose a new state $x' \sim q(x' \mid x)$\\
    Calculate the acceptance ratio $\alpha = \min\left(1, \frac{p(x') q(x \mid x')}{p(x) q(x' \mid x)}\right)$\\
    Draw a uniform random number $u \sim \text{Uniform}(0,1)$\\
    \eIf{$u \leq \alpha$}{
        Accept the new state $x'$\\
        $x = x'$
    }{
        Reject the new state and retain the current state $x$
    }
    Record the state $x$
}
\end{algorithm}

\subsubsection{Degree Corrected SBM}

The main idea behind the Degree Corrected SBM (DC-SBM) \cite{funke2019stochastic} is the introduction of a new parameter $\theta = \{\theta_1, \ldots, \theta_N\}$, which controls the expected degree of each node. This parameter allows modeling the heterogeneity of degrees within groups.

\begin{figure}[H]
    \centering
    \includegraphics[width=\linewidth]{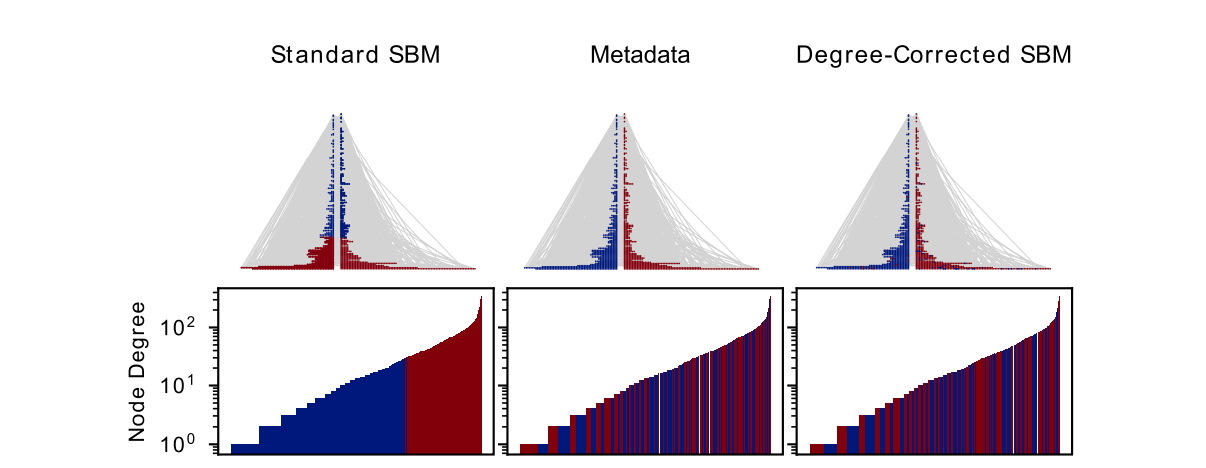}
    \caption{Comparison between Standard SBM and DC-SBM: Standard SBM groups nodes based on degree similarity, while DC-SBM allows for a more varied degree distribution within communities \cite{funke2019stochastic}}
    \label{fig:sbmdcsbm}
\end{figure}

In the standard SBM, the probability of having an edge between two nodes $i$ and $j$ belonging to groups $r$ and $s$, respectively, is given by:
\[
P(i \to j) = P(r \to s) = \omega_{rs}.
\]

The degree-corrected model modifies this probability to allow for degree heterogeneity. The probability of observing at least one edge between nodes $i$ and $j$ in the DC-SBM is:
\[
P(i \to j) = 1 - \exp(-\theta_i \theta_j \omega_{rs}) \text{ for } i \in b_r, j \in b_s, r \ne s.
\]
Here, $\theta_i$ and $\theta_j$ control the expected degrees of nodes $i$ and $j$, respectively.
Examining Figure \ref{fig:sbmdcsbm}, it is clear that within a bloc, nodes can possess different degrees. This illustrates the adaptability of DC-SBM, which accommodates diverse degree distributions within communities, contrasting with the stricter grouping of nodes based solely on degree similarity in Standard SBM.

\paragraph{Heterogeneity of Degrees}

The parameters $\theta_i$ allow each node to have a different expected degree, even if it belongs to the same block. This means that DC-SBM can better represent communities where some nodes have many connections (high degrees) and others have few (low degrees).

\paragraph{Likelihood Function}

The likelihood function formulated by authors for the DC-SBM is:
\[
L_{KN} = \sum_{rs} e_{rs} \log \left( \frac{e_{rs}}{e_r e_s} \right)
\]
where $e_{rs}$ is the number of edges between blocks $r$ and $s$, and $e_r = \sum_s e_{rs}$ is the total number of edges connecting nodes in a block $r$.

The microcanonical variant of the DC-SBM, according to Peixoto \cite{karrer2011stochastic}, is approximately:
\[
L_P \approx M + \sum_k N_k \log(k!) + \frac{1}{2} \sum_{rs} e_{rs} \log \left( \frac{e_{rs}}{e_r e_s} \right)
\]
where $N_k$ is the number of nodes with degree $k$. This formulation differs slightly from that of Karrer and Newman in terms of constant and multiplicative terms, but these differences do not affect the optima for a fixed number of groups. However, they can influence the model selection methods.

\paragraph{Advantages}

The introduction of the parameter $\theta$ allows for a broader distribution of degrees within inferred groups, which better matches the characteristics of real-world networks. For example, in the standard SBM, nodes with similar degrees tend to be grouped, while the DC-SBM allows for a more varied distribution of degrees within groups.

In the end, the DC-SBM significantly improves the standard SBM by allowing for better modeling of real-world networks with broad degree distributions. This is particularly useful for scale-free graphs, where node degrees can vary significantly.

\subsection{Markov Clustering Algorithm}
Markov Clustering (MCL), from \cite{Van_Dongen_2008}, is an efficient and scalable algorithm used to cluster graphs. It leverages the concept of random walks and the properties of Markov chains to identify densely connected regions, or clusters, within a graph. The algorithm operates in two main phases: expansion, which simulates the spread of a random walk to model the natural tendency of nodes to cluster together, and inflation, which sharpens these clusters by amplifying stronger connections while weakening weaker ones. MCL is particularly well-suited for large and complex networks, making it a popular choice for tasks such as protein interaction network analysis, social network analysis, and other applications where discovering community structure is essential.

\subsubsection{Random Walks and Markov Chains}
A \textit{random walk} is a mathematical formalization of a path consisting of a succession of random steps. It is a stochastic process that describes a sequence of possible movements in a mathematical space, where the next position is determined randomly from the current position.

Taking a graph \(G = (V, E)\), a random walk on \(G\) starts at an initial vertex \(v_0 \in V\). At each step, the walk moves to a neighboring vertex connected by an edge chosen uniformly at random among all neighbors.

Let \(X_t\) denote the position of the walk at time \(t\). The process \(\{X_t\}_{t \geq 0}\) is a Markov chain. A \textit{Markov Chain} is a stochastic process that satisfies the Markov property, which states that the future state depends only on the current state and not on the previous sequence of events. Formally, this can be expressed as:
\[
P(X_{t+1} = x \mid X_t = x_t, X_{t-1} = x_{t-1}, \ldots, X_0 = x_0)
\]
\[
 = P(X_{t+1} = x \mid X_t = x_t).
\]

The possible states of the system are represented as vertices in a graph, and the transitions between states are represented as edges with associated uniform probabilities.

The transition probabilities of a Markov chain can be represented in a \textit{transition matrix} \(\mathbf{P}\), where each element \(P_{uv}\) indicates the probability of transitioning from state (or vertex) \(u\) to state \(v\). For a random walk on a graph \(G\), the transition probability is given by:
\begin{equation} \label{eq:probmat}
P_{uv} = 
\begin{cases} 
\frac{1}{\deg(u)} & \text{if } (u, v) \in E \\
0 & \text{otherwise}
\end{cases}
\end{equation}

The matrix \(\mathbf{P}\) is a square matrix of size \( |V| \times |V| \) and is a stochastic matrix, meaning that the sum of the elements in each row is 1. This reflects the fact that from any given vertex, the random walk must move to one of its neighboring vertices in the next step.

\subsubsection{Markov Clustering Iterative Process}

\paragraph{Expansion Step}
The expansion step involves taking the power of the transition matrix to simulate the effect of multiple steps of the random walk. Mathematically, this can be represented as:
\[
    \mathbf{P^{\text{(exp)}}} = \mathbf{P}^e
\]
where \(e \geq 2\) is the expansion parameter. This step increases the connectivity of the graph, allowing the random walk to explore more distant vertices.

\paragraph{Inflation Step}
The inflation step is used to strengthen the probabilities of intra-cluster edges while weakening inter-cluster edges. This is achieved by raising each element of the matrix to a power \(r\) and then normalizing each column. The process can be expressed as:
\[
P^{\text{(inf)}}_{ij} = \frac{(P^{\text{(exp)}}_{ij})^r}{\sum_{k} (P^{\text{(exp)}}_{kj})^r}
\]
where \(r > 1\) is the inflation parameter. This step amplifies the probabilities of stronger connections and diminishes weaker ones, emphasizing the clusters.

\begin{figure}[H]
    \centering
    \includegraphics[width=\linewidth]{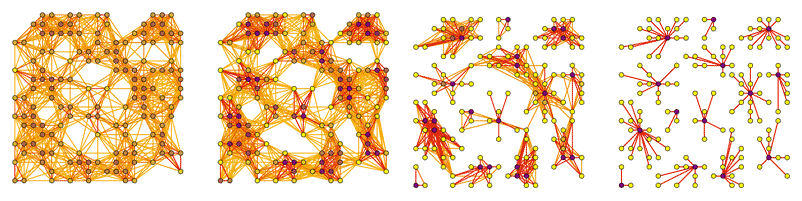}
    \caption{Illustration of three iterations of the Markov Clustering Algorithm \cite{Van_Dongen_2008}}
    \label{fig:markov}
\end{figure}

The expansion and inflation steps are repeated iteratively until convergence (see Figure \ref{fig:markov}). Convergence is typically achieved when the matrix stabilizes and does not change significantly between iterations.

\paragraph{Find the clusters} In the end, vertices that are strongly connected in $\mathbf{P}$ (i.e. have high transition probabilities) form clusters. Clusters are found by considering the non-zero elements of the diagonal as \textit{attractors} (cluster centers). All nodes which have non-zero values in an attractor row form a cluster. Algorithm \ref{alg:markov} explains the complete clustering process.

\begin{algorithm}[H]
    \caption{Markov Clustering Algorithm}
    \label{alg:markov}
    \SetAlgoLined
    \KwIn{Graph $G = (V, E)$ with adjacency matrix $\mathbf{A}$, expansion parameter $e$, inflation parameter $r$, threshold $\epsilon$}
    \KwOut{Cluster assignments $\mathcal{C}=(C_1, C_2, \dots, C_k)$}
    \BlankLine
    Get the probability matrix $\mathbf{P}$ using Eq. \eqref{eq:probmat}\\
    \Do{$\max_{i,j}\lvert P_{ij} - P^{\text{\normalfont (prev)}}_{ij}\rvert > \epsilon$}{
        Let $\mathbf{P}^{\text{(prev)}} = \mathbf{P}$\\
        \BlankLine
        $\mathbf{P} = \mathbf{P}^e$ \tcp*{Expansion}
        \BlankLine
        \For{$i,j = 1$ \KwTo $n$}{
                $P_{ij} = \frac{(P_{ij})^r}{\sum_k (P_{kj})^r}$ \tcp*{Inflation}
        }
    }
    Form clusters $\mathcal{C} = (C_1, C_2, \dots, C_k)$ by grouping nodes corresponding to the rows of $\mathbf{P}$ where the diagonal elements are non-zero
\end{algorithm}

\subsection{Leiden algorithm}

In 2019, \cite{leiden} proposed a new method based on the Louvain algorithm \cite{louvain}, demonstrating significantly improved performance over the original algorithm \cite{leiden, louvainleidercomp}. 

Like its predecessor, it is based on the optimization of a quality metric. However, to overcome some limitations of modularity, the authors used another quality metric, called the Constant Potts Model (CPM) \cite{Traag_van_Dooren_Nesterov_2011}, defined as:
\begin{equation} \label{eq:cpm}
    \mathcal{H} = \sum_{C \in \mathcal{C}} \left[ \lvert E(C,C) \rvert - \gamma \binom{n_c}{2} \right]
\end{equation}

where $\lvert E(C,C)\rvert$ is the number of inter-cluster edges in $C$, $n_c$ the number of nodes in the cluster $C$, and $\gamma$ a hyperparameter defining the clusters density.

\begin{figure}[H]
    \centering
    \includegraphics[width=\linewidth]{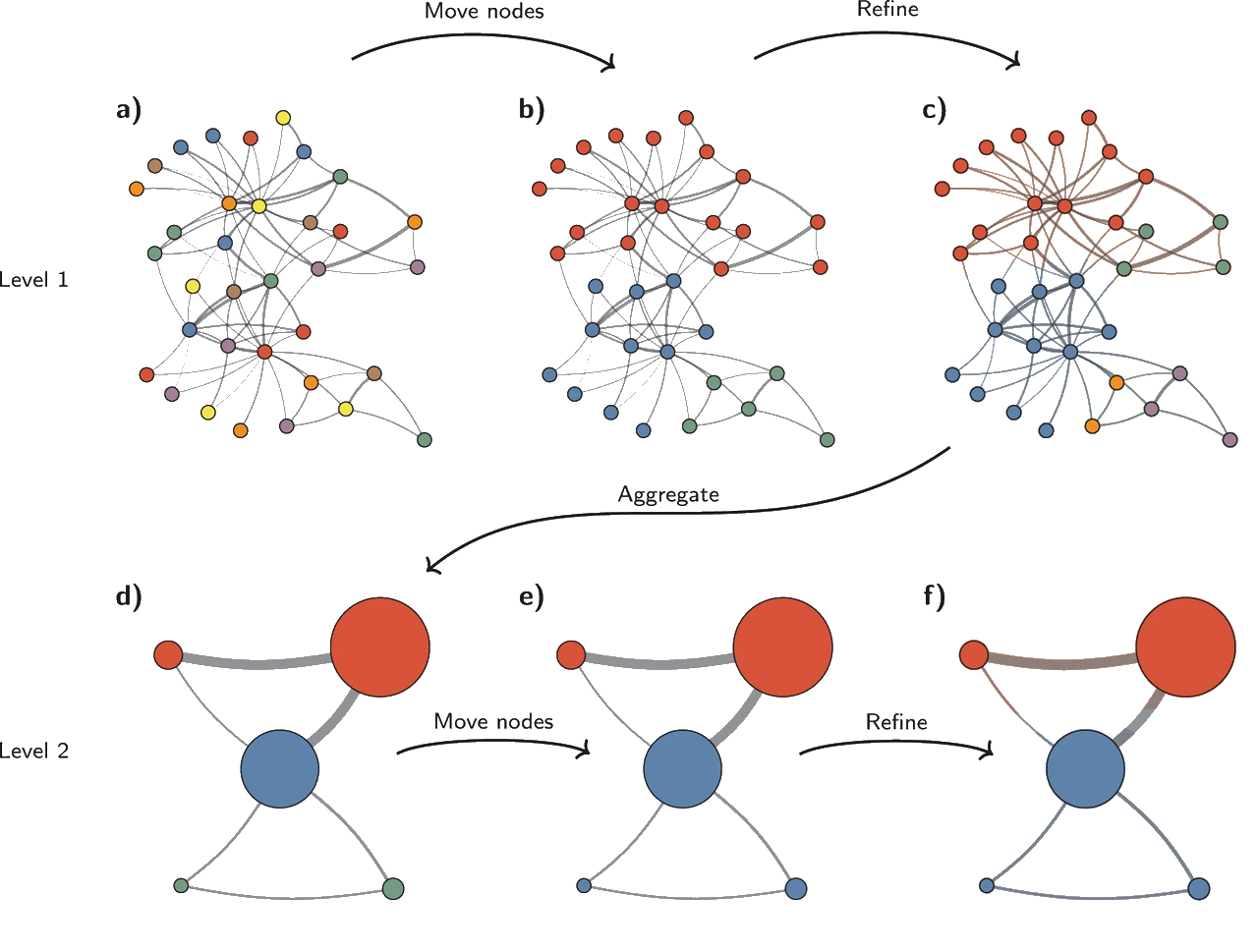}
    \caption{Leiden algorithm \cite{leiden}}
    \label{fig:leiden}
\end{figure}

The main issue of the Louvain algorithm is the case of “weakly--connected” communities. Indeed, a community may be internally 'disconnected' but remains as one community, the Leiden algorithm does not have this issue. The Leiden algorithm follows the same general phases as Louvain with an additional phase \cite{leiden}. The algorithm is illustrated in Figure \ref{fig:leiden}

\paragraph{Local nodes movement} The algorithm starts where each node of the graph $G$ forms a cluster. Next, by optimizing an objective function such as modularity, individual clusters are moved from one community to another to form a partition $P$ of the graph. 

\paragraph{Refinement} The main idea behind the refinement phase is to identify sub-communities in $P$ to form a refined partition $P_r$. At the beginning of the Leiden algorithm $P_r = P$, then the algorithm locally merges nodes in $P_r$, the important fact is that mergers are made only within each community of $P$. A node is merged with a community in $P_r$ only if both are well-connected to their community in $P$. After the refinement phase, the communities in $P$ are often split into multiple communities in $P_r$.

\paragraph{Network aggregation} An aggregate network is created based on the refined partition $P_r$, using the non-refined partition $P$ to create an initial partition for the aggregate network. Nodes and edges of the aggregate network represent communities and the relationships between them in the refined partition $P_r$.

\paragraph{Iteration} The previous phases are repeated until there are no further changes in the partition, i.e., until the partition is stable.

The summarized algorithm is presented in Algorithm \ref{alg:leiden}

\begin{algorithm}[H]
    \caption{Leiden Algorithm}
    \label{alg:leiden}
    \SetAlgoLined
    \KwIn{Graph $G = (V, E)$}
    \KwOut{Cluster assignments $\mathcal{C}=(C_1, C_2, \dots, C_k)$}
    \BlankLine
    \Do{changes in partition during the iteration}{
        For each node in $G$, the nodes move to the community that maximizes the modularity and form the partition $P$\\
        The partition $P$ is refined to form $P_r$ \\
        Aggregate the network based on the refined partition $P_r$ \\
        Form cluster assignments $\mathcal{C}=(C_1, C_2, \dots, C_k) $ thanks to the aggregated network.
    }
    
\end{algorithm}

\section{Deep Graph Clustering} \label{sec:deep}

Recent advances in Deep Learning and especially Graph Neural Networks brought a new field of study into Graph Clustering. The high representation capabilities of GNNs are, indeed, what is sought in methods such as Spectral Clustering (see section \ref{ssec:spectral}), which creates another representation of the graph via spectral decomposition. These methods exploit the capacity of Neural Networks to uncover non-linear intricate patterns and data structure within the graph. The rise in interest in this paradigm has sparked a surge in innovation, resulting in the development of several approaches and models, as seen in the different surveys \cite{Wang_Yang_Yao_Bai_Zhu_2024, Liu_Xue_Wu_Zhou_Hu_Paris_Nepal_Yang_Yu_2020, Su_Xue_Liu_Wu_Yang_Zhou_Hu_Paris_Nepal_Jin_et_al_2024, deep_graph_clustering_survey}.

\subsection{Overview}

\begin{figure}[H]
    \centering
    \includegraphics[width=.92\linewidth]{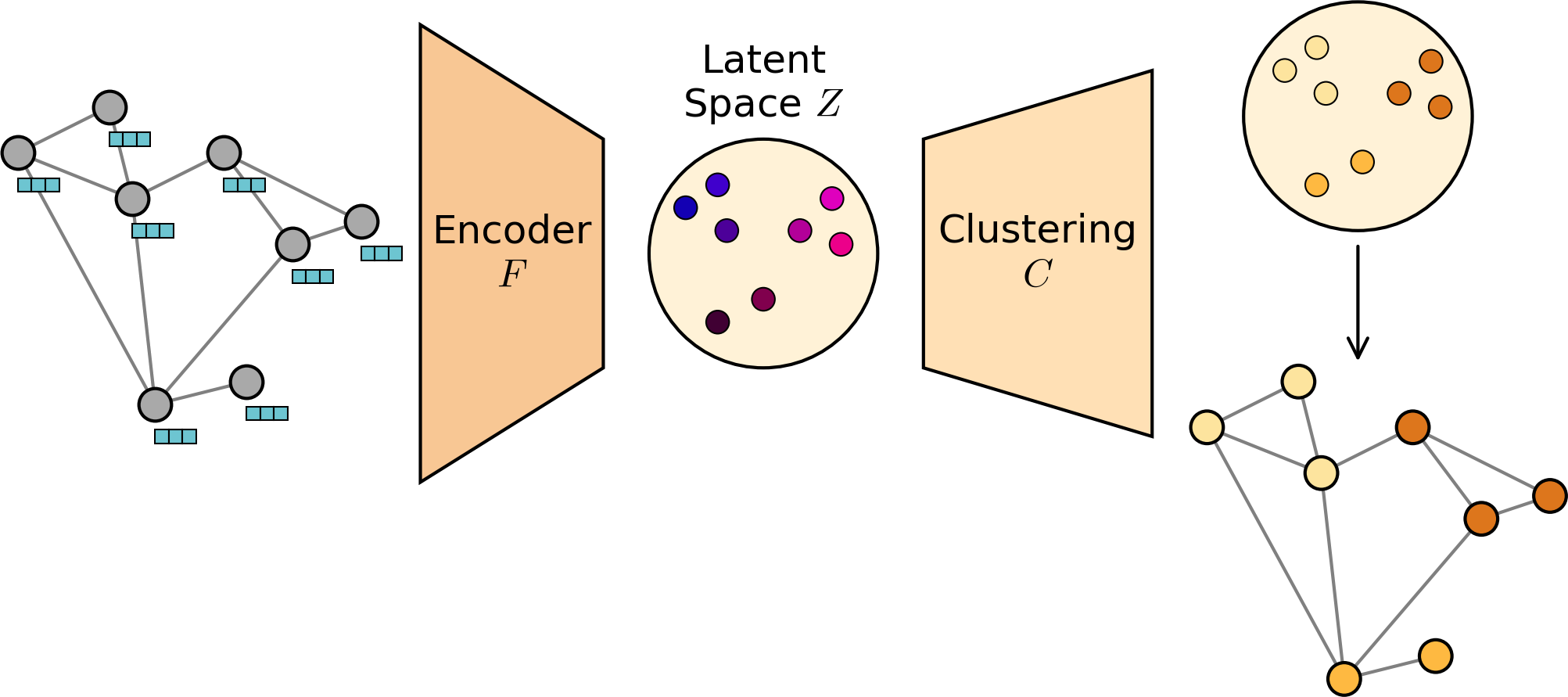}
    \caption{Deep Graph Clustering framework}
    \label{fig:deep-framework}
\end{figure}

Almost all deep-graph clustering (DGC) methods lie on a similar framework. As illustrated in Figure \ref{fig:deep-framework}, the graph is given to a neural network $F$, whose task is to encode the nodes into a latent continuous space $\mathbf{Z}$ using the graph structure and the node features. A traditional clustering technique $C$, such as $k$-means, is then used on this latent space to determine the different clusters.

Three main Deep Graph Clustering learning approaches are highlighted in \cite{deep_graph_clustering_survey}. The first is the \textit{reconstructive} approach, which focuses on reconstructing the structure of the graph. The second is the \textit{adversarial} approach, inspired by the success of Generative Adversarial Networks \cite{goodfellow_generative_2014}, which aims to align the latent space with a specific distribution using adversarial techniques. The third is the \textit{contrastive} approach, which trains the model to differentiate between similar and dissimilar entities. This section presents a study of three models representing each learning method.

\subsection{Graph Autoencoder} \label{ssec:gae}

\begin{figure}[H]
    \centering
    \includegraphics[width=\linewidth]{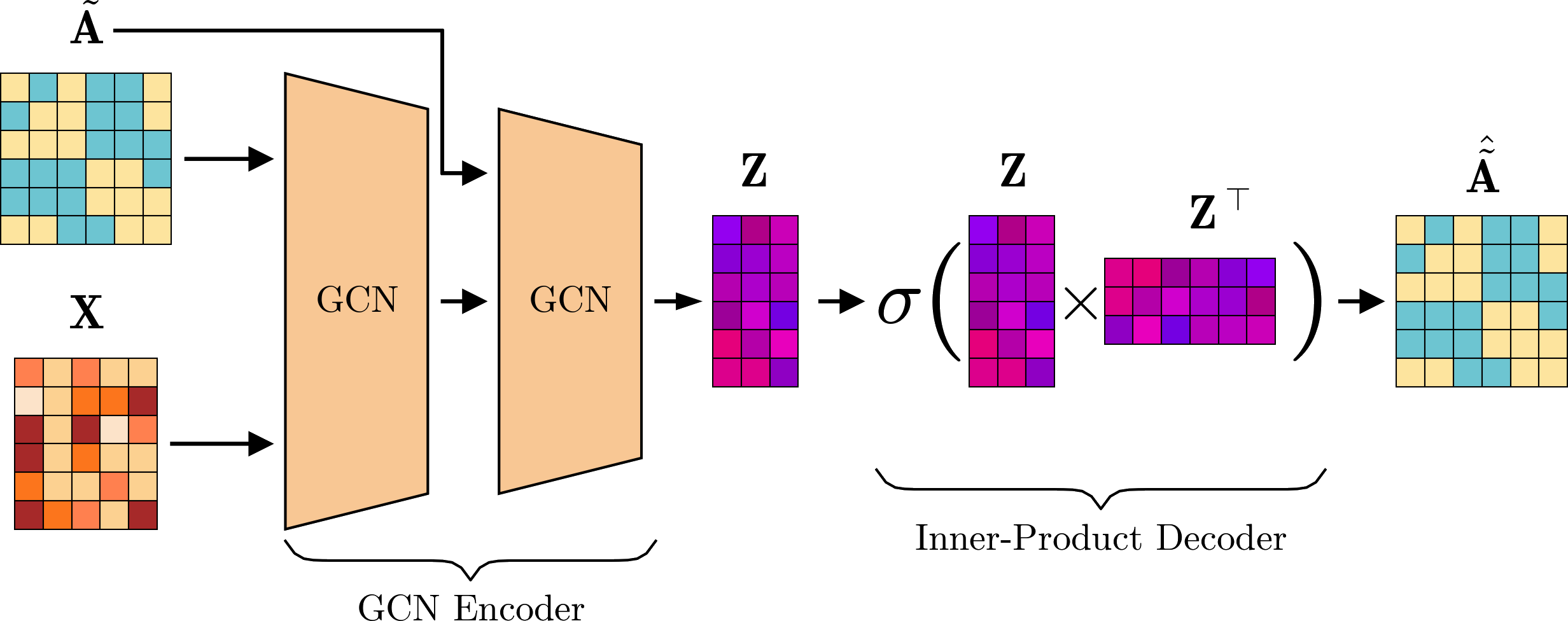}
    \caption{Graph Autoencoder architecture}
    \label{fig:gae}
\end{figure}

The first significant contribution to Deep Graph Clustering was introduced in 2016 with the Graph Autoencoder (GAE) by \cite{Kipf_Welling_2016}. Although the original paper focused on link prediction, the model has been widely adopted for Graph Clustering. The GAE employs an autoencoder architecture \cite{Hinton2006} coupled with GCNs \cite{Kipf_Welling_2017} to encode graph nodes into a latent space and then reconstruct the original adjacency matrix. Graph convolutional layers are used in the encoder to incorporate the graph structure in the encoding process, while the decoder reconstructs the adjacency matrix by taking the inner product of the latent matrix and applying a sigmoid function. The model architecture is illustrated in Figure \ref{fig:gae}. It illustrates the \textit{reconstructive} approach, as it aims to reconstruct the graph. 

Let $G = (V, E)$ be a graph with $n = |V|$ nodes, and let $\mathbf{X}$ be the features matrix of shape $n \times d$, where \( d \) is the number of features per node. The adjacency matrix of \( G \) is denoted as $\mathbf{A}$. The encoder consists of several GCN layers, typically two. The latent encoded matrix $\mathbf{Z}$ of the node features of size $n \times d_h$, where $d_h$ is the latent dimension, and where each row represents the corresponding node of the original feature matrix, is calculated as follows:
\[
\mathbf{Z} = \text{GCN}(\mathbf{X}, \mathbf{A})
\]
The adjacency matrix is then reconstructed using an inner product followed by a sigmoid activation:
\[
    \hat{\mathbf{A}} = \sigma \left( \mathbf{Z} \mathbf{Z}^{\top} \right)
\]
The model is trained to reconstruct the adjacency matrix, encouraging the latent space to integrate the graph structure into its transformation. Using an inner product to reconstruct the adjacency matrix is particularly effective, as it leverages the cosine similarity of two rows -- the latent representations of nodes -- to infer the similarity, and thus the connection, between two nodes.

For clustering, training is performed on the entire graph for \( e \) iterations (epochs). During each forward pass, the model's loss, referred to as \textit{reconstruction loss} \(\mathcal{L}_{\text{enc}}\), is computed by comparing the decoder's output adjacency matrix to the actual adjacency matrix using the Binary Cross-Entropy Loss:
\begin{equation} \label{eq:bceloss}
    \mathcal{L}_{\text{enc}} = 
        \frac{-1}{n^2}\sum_{i=1}^n\sum_{j=1}^n \left(
            A_{ij}\log \hat{A}_{ij} +
            (1 - A_{ij})\log (1 - \hat{A}_{ij})
        \right)
\end{equation}

Once the model is trained, it can be used for clustering. This is achieved by encoding the graph with the
GCN encoder of the GAE, resulting in the latent matrix $\mathbf{Z}$, where each row corresponds to a node. A traditional clustering algorithm, such as $k$-means, is then applied to $\mathbf{Z}$. The output of this algorithm will be the clustered nodes. Algorithm \ref{alg:gae} presents the complete clustering process with GAE.

\begin{algorithm}[H]
\caption{Clustering with Graph Autoencoder}
\label{alg:gae}
\SetAlgoLined
\KwIn{Graph $G = (V, E)$ with adjacency matrix $\mathbf{A}$ and feature matrix $\mathbf{X}$, number of epochs $e$, number of clusters $k$}
\KwOut{Cluster assignments $\mathcal{C}=(C_1, C_2, \dots, C_k)$}
\BlankLine
\For{epoch $= 1$, $2$, $\cdots$,  $e$}{
    Perform a forward pass through the GCN encoder to get the latent representation $\mathbf{Z}$\\
    Reconstruct the adjacency matrix $\hat{\mathbf{A}} = \sigma(\mathbf{Z} \mathbf{Z}^{\top})$\\
    Calculate the loss $\mathcal{L}_{\text{enc}}$ using Eq. \eqref{eq:bceloss}\\
    Backpropagate the loss $\mathcal{L}_{\text{enc}}$ and update the model parameters
}
Encode the graph to obtain the latent representation $\mathbf{Z}$:
\[
    \mathbf{Z} = \text{GCN}(\mathbf{X}, \mathbf{A})
\]\\
Apply \( k \)-means clustering in \(\mathbf{Z}\):
\[
\mathcal{C} = k\text{-means}(\mathbf{Z}, k)
\]
\end{algorithm}

\subsection{Adversarially Regularized Graph Autoencoder}

\begin{figure}[H]
    \centering
    \includegraphics[width=\linewidth]{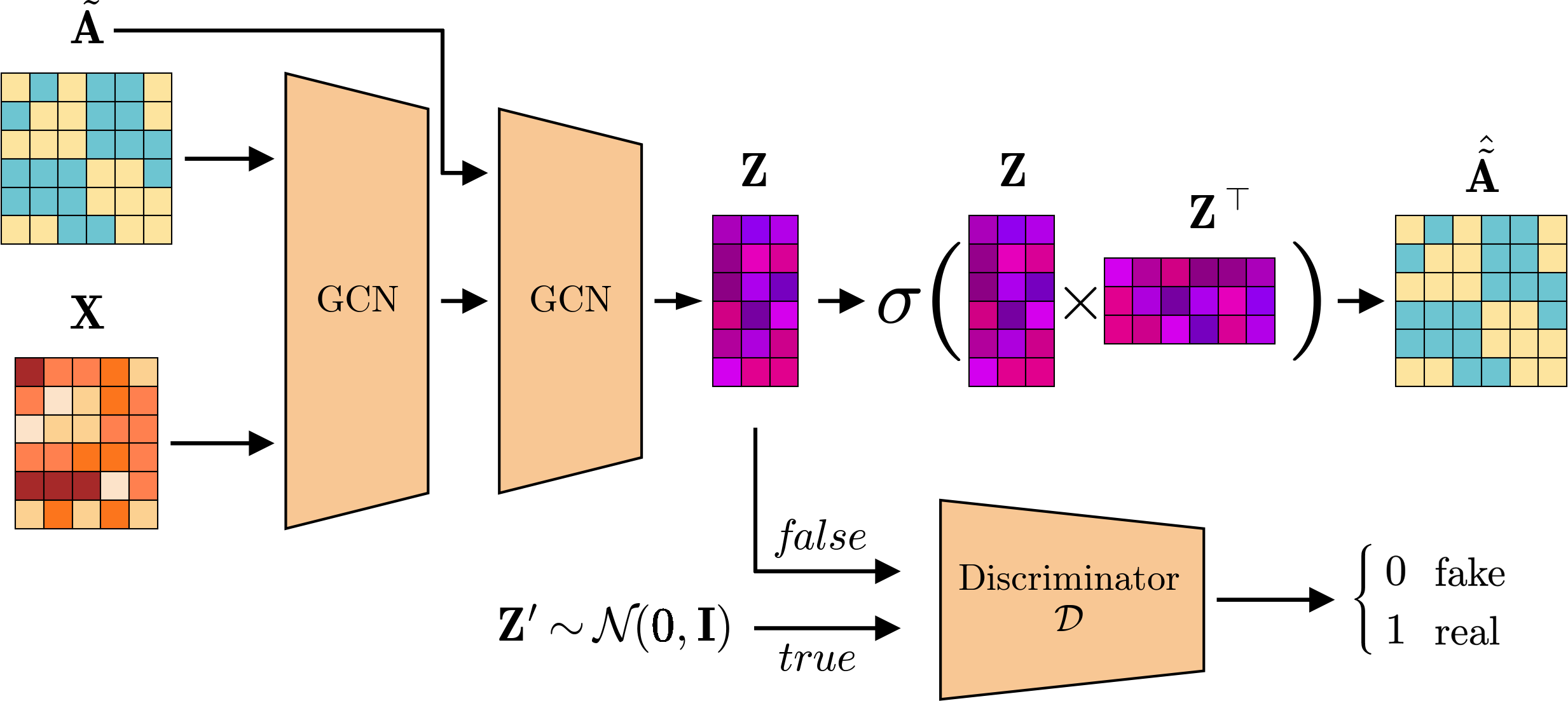}
    \caption{Adversarially Regularized Graph Autoencoder architecture}
    \label{fig:arga}
\end{figure}

Motivated by the advancements in generative Deep Learning, especially Generative Adversarial Neural Networks (GANs) \cite{goodfellow_generative_2014}, \cite{Pan_Hu_Long_Jiang_Yao_Zhang_2018} proposed a novel architecture based on the Graph Autoencoder (GAE), the Adversarially Regularized Graph Autoencoder (ARGA), incorporating an adversarial mechanism for latent space regularization. Figure \ref{fig:arga} illustrates this model. It represents the \textit{adversarial} learning approach.

This method employs the same GAE model as described in \ref{ssec:gae}, but with a crucial difference in the update strategy. The authors introduce an adversarial technique where the latent node representations are compared to randomly generated vectors (sampled from a Gaussian distribution) through a discriminator neural network. In this setup, the random matrix is labeled as true, while the real matrix is labeled as false. The encoder and the discriminator are indirectly competing with each other: the encoder aims to fool the discriminator into classifying the real matrix as from the Gaussian distribution, thereby enhancing the regularization of the latent space, while the discriminator tries to accurately distinguish between the real and random matrices.

Let \( G = (V, E) \) be a graph with \( n = \lvert V \rvert \) nodes and node features \( \mathbf{X} \). The adjacency matrix of \( G \) is denoted by \( \mathbf{A} \). The ARGA uses an encoder \( \text{GCN}(\cdot, \cdot) \), where \( \mathbf{Z} = \text{GCN}(\mathbf{X}, \mathbf{A}) \) is the latent encoded matrix of shape \( n \times d_h \). Furthermore, the model employs a discriminator \( \mathcal{D} \), a dense neural network that takes input vectors of shape \( d_h \) and outputs a single value for each vector, followed by a sigmoid activation. This output represents the probability, in the range \( ]0,1[ \), that the given vector is a real vector from the GCN encoder and not a randomly sampled vector \( \mathbf{z'} \sim \mathcal{N}(0, 1) \).

During the training of the model, the representation of the latent node \( \mathbf{Z} = \text{GCN}(\mathbf{X}, \mathbf{A}) \) is calculated, and the random vector of shape $n$ \( \mathbf{z_i'} \sim \mathcal{N}(0,1) \) is sampled, forming the matrix $\mathbf{Z}'$. Both matrices \( \mathbf{Z} \) and \( \mathbf{Z}' \) are fed into the discriminator \( \mathcal{D} \). The encoder is trained to reconstruct the adjacency matrix through the decoder \( \sigma(\mathbf{Z}\mathbf{Z}^\top) \), similar to the GAE, while the discriminator is trained to accurately classify the input as true latent node representations or randomly sampled vectors.

Two loss functions are used for this model. The first being the reconstruction loss $\mathcal{L}_\text{enc}$, from Eq. \eqref{eq:bceloss} for the encoder, and the second being $\mathcal{L}_\text{disc}$, also using the Binary Cross Entropy Loss:
\begin{equation} \label{eq:discloss}    
    \mathcal{L}_\text{disc}=-\frac{1}{n}\sum_{i=1}^n\left(\log \mathcal{D}(\mathbf{z'_i}) + \log (1-\mathcal{D}(\mathbf{z_i})) \right)
\end{equation}

This loss function is used at the same time in the discriminator and in the encoder, enabling the encoder to approach a Gaussian distribution in its latent space, therefore creating a regularized and coherent space.

\begin{algorithm}[H]
\caption{Clustering with Adversarially Regularized Graph Autoencoder}
\label{alg:arga}
\SetAlgoLined
\KwIn{Graph $G = (V, E)$ with adjacency matrix $\mathbf{A}$ and feature matrix $\mathbf{X}$, number of epochs $e$, discriminator training iterations $K$, number of clusters $k$}
\KwOut{Cluster assignments $\mathcal{C}=(C_1,C_2,\dots,C_k)$}
\BlankLine
\For{epoch $= 1$, $2$, $\cdots$,  $e$}{
    Perform a forward pass through the GCN encoder to get the latent representation $\mathbf{Z}$\\
    \For{$i=1$, $2$, $\cdots$, $K$}{
        Sample $\mathbf{Z'}\sim \mathcal{N}(0,1)$ of shape $n\times d_h$\\
        Compute $\mathcal{D}$ loss $\mathcal{L}_\text{disc}$ via Eq. \eqref{eq:discloss}\\
        Backpropagate the loss $\mathcal{L}_\text{disc}$ and update the discriminator model parameters
    }
    Reconstruct the adjacency matrix $\hat{\mathbf{A}} = \sigma(\mathbf{Z} \mathbf{Z}^{\top})$\\
    Compute the encoder loss $\mathcal{L}_{\text{enc}}$ using Eq. \eqref{eq:bceloss}\\
    Backpropagate the loss $\mathcal{L}_{\text{enc}}$ and update the encoder and encoder model parameters
}
Encode the graph to obtain the latent representation $\mathbf{Z}$:
\[
\mathbf{Z} = \text{GCN}(\mathbf{X}, \mathbf{A})
\]\\
Apply \( k \)-means clustering in \(\mathbf{Z}\):
\[
\mathcal{C} = k\text{-means}(\mathbf{Z}, k)
\]
\end{algorithm} 

The clustering is also done using the latent matrix $\mathbf{Z}$, such as described in the GAE section. The algorithm \ref{alg:arga} explains the clustering method with ARGA.

\subsection{Multi-View Graph Representation Learning}

\begin{figure}[H]
    \centering
    \includegraphics[width=\linewidth]{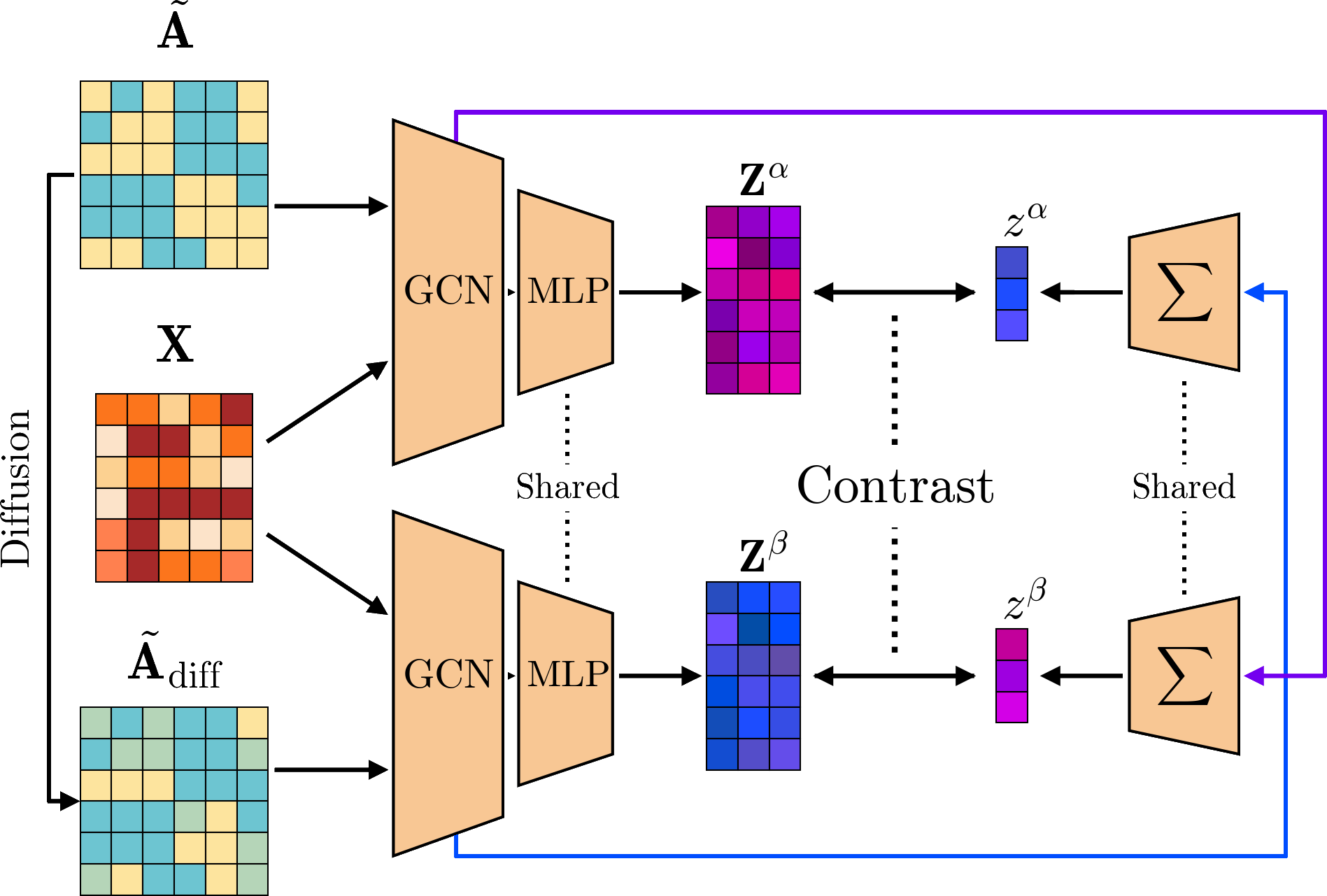}
    \caption{Multi-View Graph Representation Learning architecture ($\sum$ illustrates the graph pooling, and the contrast should maximize the MI)}
    \label{fig:mvgrl}
\end{figure}

The third learning approach, called \textit{contrastive}, has been used by \cite{Hassani_Khasahmadi_2020} for the Multi-View Graph Representation Learning (MVGRL) model. This approach takes advantage of recent advances in contrastive learning \cite{Li_Gu_Dullien_Vinyals_Kohli_2019}, an unsupervised learning paradigm in which data points are compared to one another, allowing the model to discern similar and different data points. Additionally, they use multi-view representation learning \cite{Tian_Krishnan_Isola_2020,Bachman2019}, which is a data augmentation technique used to generate multiple views of the same data for contrastive learning. The main contribution is the model's objective, which is to maximize the Mutual Information (MI) between representations encoded from different views of the graph. This allows the model to learn meaningful representations that capture similarities and differences within the graph. The architecture and learning method is illustrated in Figure \ref{fig:mvgrl}. 

The first step of the method is to create another view of the graph. The authors argue that two views are sufficient and that more views do not lead to better results. To generate this alternative representation, graph diffusion, particularly Personalized PageRank (PPR) \cite{Page_Brin_Motwani_Winograd_1999}, is used. This technique models how a node can spread through a network. For a graph with an adjacency matrix $\mathbf{A}$ of shape \( n \times n \) and a degree matrix $\mathbf{D}$, it is defined as:
\begin{equation} \label{eq:ppr}
    \mathbf{S} = \alpha \left( \mathbf{I}_n - (1 - \alpha) \mathbf{D}^{-\sfrac{1}{2}} \mathbf{A} \mathbf{D}^{-\sfrac{1}{2}} \right)^{-1}
\end{equation}
where $\alpha$ is a parameter and $\mathbf{S}$ is the resulting diffusion matrix, representing the adjacency matrix of the second view.

The graphs are then fed to two GCNs, each one composed of one or more layers. Finally, the outputs of the GCNs are given to a shared MLP, the projection head. The results are the encoded node representations $\mathbf{Z}^\alpha$ and $\mathbf{Z}^\beta$ of the original and diffused graph, respectively. These operations can be written as:
\begin{align} \label{eq:noderep}
    \begin{aligned}
        \mathbf{Z}^\alpha&=\text{MLP}(\text{GCN}(\mathbf{X}, \mathbf{A}))\\
        \mathbf{Z}^\beta&=\text{MLP}(\text{GCN}(\mathbf{X}, \mathbf{S}))
    \end{aligned}
\end{align}
Note that the MLP is shared between the two encoders.

To apply a contrastive mechanism, the node representations learned from the GCNs are utilized in a pooling method. The average output of each GCN layer, taken across all nodes for each dimension, is concatenated and fed into a single-layer neural network. This pooling method is defined as:
\begin{equation} \label{eq:pooling}
    \mathbf{z_g} = \sigma\left(\concat_{l=1}^{L} \left[ \frac{1}{n}\sum_{i=1}^n \mathbf{z_i^{(l)}} \right] \mathbf{W} \right)
\end{equation}

where $\mathbf{z_g} \in \mathbb{R}^{d_h}$ is the encoded graph representation, $L$ is the number of GCN layers, $\mathbf{z_i^{(l)}}$ is the output of the $i$-th node in the $l$-th GCN layer, $\concat$ is the concatenation operator, $\mathbf{W} \in \mathbb{R}^{(L \times d_h) \times d_h}$ are the network parameters, and $\sigma$ is an activation function. This pooling method results in the two final representations for the two graph views: $\mathbf{z^\alpha}$ and $\mathbf{z^\beta}$.

During the training phase, the model aims to maximize the MI between $\mathbf{Z}^\alpha$ and $\mathbf{z^\beta}$, as well as between $\mathbf{Z}^\beta$ and $\mathbf{z^\alpha}$. To achieve this, a discriminator $\mathcal{D}$ is used to score the agreement between the two representations. In the implementation, a bilinear layer followed by a sigmoid activation is employed, which is defined as $y = \sigma(\mathbf{x_1}^\top \mathbf{W} \mathbf{x_2} + \mathbf{b})$.

For training, the discriminator is provided positive samples (real nodes and graph representations, labeled \textit{true}) and negative samples (a real graph representation and fake node representations, labeled \textit{false}). The networks aim to distinguish between these two scenarios. This leads to the maximization of MI between the two real representations during the encoding.
To generate fake node representations, the graph features $\mathbf{X}$ are shuffled into
$\mathbf{X}_c$ and fed through the nodes encoder. These corrupted representations are denoted
$\mathbf{Z}^\alpha_c$ and $\mathbf{Z}^\beta_c$. The loss function $\mathcal{L}$ uses the Binary Cross-Entropy loss and is therefore defined as:
\begin{align} \label{eq:bce_mvgrl}
\begin{aligned}
    \mathcal{L}=&\frac{1}{n}\sum_{i=1}^n(\log \mathcal{D}(\mathbf{Z}^\alpha, \mathbf{z^\beta})+\log (1-\mathcal{D}(\mathbf{Z}^\alpha_c, \mathbf{z^\beta}))\\
    &+\log \mathcal{D}(\mathbf{Z}^\beta, \mathbf{z^\alpha})+\log (1-\mathcal{D}(\mathbf{Z}^\beta_c, \mathbf{z^\alpha})))
\end{aligned}
\end{align}

After training the model, the final node representation is obtained through $\mathbf{Z}=\mathbf{Z}^\alpha+\mathbf{Z}^\beta$, and the final graph representation with $\mathbf{z}=\mathbf{z^\alpha} + \mathbf{z^\beta}$. The clusters are obtained in the same way as the other learning methods. Indeed, $\mathbf{Z}$ is used in a standard algorithm to provide the clusters $\mathcal{C}$. The complete process is illustrated in Algorithm \ref{alg:mvgrl}.

\begin{algorithm}[H]
\caption{Clustering with Multi-View Graph Representation Learning}
\label{alg:mvgrl}
\SetAlgoLined
\KwIn{Graph $G = (V, E)$ with adjacency matrix $\mathbf{A}$ and feature matrix $\mathbf{X}$, number of epochs $e$, PPR parameter $\alpha$, number of clusters $k$}
\KwOut{Cluster assignments $\mathcal{C}=(C_1,C_2,\dots,C_k)$}
\BlankLine
Compute the diffusion matrix $\mathbf{S}$ via Eq. \eqref{eq:ppr}\\
Obtain the corrupted features $\mathbf{X}_c$ by shuffling the node features $\mathbf{X}$\\
\For{epoch $= 1$, $2$, $\cdots$,  $e$}{
    Compute the encoded node representations $\mathbf{Z}^\alpha$ and $\mathbf{Z}^\beta$ with Eq. \eqref{eq:noderep}\\
    Compute the negative samples $\mathbf{Z}^\alpha_c$ and $\mathbf{Z}^\beta_c$ with $\mathbf{X}_c$ and Eq. \eqref{eq:noderep} \\
    Compute the encoded graphs representations $\mathbf{z^\alpha}$ and $\mathbf{z^\beta}$ via Eq. \eqref{eq:pooling}\\
    Compute the loss of the model $\mathcal{L}$ using Eq. \eqref{eq:bce_mvgrl}\\
    Backpropagate the loss $\mathcal{L}$ and update the GCN, MLP and discriminator parameters
}
Encode the nodes with Eq. \eqref{eq:noderep} to obtain the latent representation $\mathbf{Z}=\mathbf{Z}^\alpha+\mathbf{Z}^\beta$\\
Apply \( k \)-means clustering on \(\mathbf{Z}\):
\[
\mathcal{C} = k\text{-means}(\mathbf{Z}, k)
\]
\end{algorithm}

\begin{table*}[ht]
    \caption{Comparative experiments results of presented methods (the best results \underline{\textbf{highlighted}})}
    \centering
    \begin{tabular}{l|cccc|cccc|cccc}
        \toprule
        \multirow{2}{*}{} & \multicolumn{4}{c|}{Cora} & \multicolumn{4}{c|}{Citeseer} & \multicolumn{4}{c}{UAT}\\
        \cmidrule{2-13}
        & ACC & NMI & ARI & Q & ACC & NMI & ARI & Q & ACC & NMI & ARI & Q \\
        \midrule
        \multicolumn{13}{c}{Traditional Clustering}\\
        \midrule
        Spectral \cite{von_Luxburg_2007} & 30.2\% & 0.013 & 0.007 & 0.203 & 25.2\% & 0.023 & 0.018 & 0.388 & 30.8\% & 0.016 & 0.013 & 0.114 \\
        SBM \cite{Lee_Wilkinson_2019} & 30.2\% & 0.027 & 0.013 & 0.182 & 29.6\% & 0.040 & 0.029 & 0.610 & 47.6\% & 0.201 & 0.158 & 0.078 \\
        Markov \cite{Van_Dongen_2008} & \textbf{\underline{89.8\%}} & 0.413 & 0.046 & 0.598 & \textbf{\underline{83.7\%}} & 0.340 & 0.012 & 0.656 & 51.6\% & 0.134 & 0.062 & 0.261 \\
        Leiden \cite{leiden} & 79.5\% & 0.470 & 0.255 & \textbf{\underline{0.823}} & 73.9\% & 0.331 & 0.102 & \textbf{\underline{0.895}} & 44.5\% & 0.106 & 0.082 & \textbf{\underline{0.332}} \\
        \midrule
        \multicolumn{13}{c}{Deep Graph Clustering}\\
        \midrule
        GAE \cite{Kipf_Welling_2016} & 73.6\% & \textbf{\underline{0.536}} & 0.530 & 0.725 & 65.0\% & 0.355 & 0.344 & 0.756 & 52.9\% & \textbf{\underline{0.290}} & \textbf{\underline{0.211}} & 0.030 \\
        ARGA \cite{Pan_Hu_Long_Jiang_Yao_Zhang_2018} & 74.6\% & 0.528 & \textbf{\underline{0.462}} & 0.714 & 65.3\% & 0.349 & 0.322 & 0.764 & \textbf{\underline{54.3\%}} & 0.270 & 0.180 & 0.080 \\
        MVGRL \cite{Hassani_Khasahmadi_2020} & 73.7\% & \textbf{\underline{0.536}} & \textbf{\underline{0.462}} & 0.659 & 68.6\% & \textbf{\underline{0.392}} & \textbf{\underline{0.383}} & 0.720 & 53.6\% & 0.256 & 0.203 & 0.044 \\
        \bottomrule
    \end{tabular}
    \label{tab:expresults}
\end{table*}

\section{Experiments \& Results} \label{sec:experiments}
\setlength{\tabcolsep}{3pt}

In this section, we conduct comparative experiments to evaluate the methods previously described. 

\subsection{Datasets}

For these experiments, we used three standard benchmark undirected graphs for graph clustering: Cora \cite{mccallum_automating_2000}, Citeseer \cite{giles_citeseer_1998}, and UAT \cite{deep_graph_clustering_survey}. The first two are citation datasets, nodes representing scientific publications, and edges representing citations. The features are based on the presence or absence of words in the articles. The latter is an aviation data set that measures airport activity. The dimensions of the data sets are given in Table \ref{tab:datasets}. 

\begin{table}[H]
    \centering
    \caption{Datasets summary}
    \label{tab:datasets}
    \begin{tabular}{l|cccc}
        \toprule
         Name & Nodes & Edges & Features & Classes\\
        \midrule 
        \textbf{Cora} & 2708 & 5429 & 1433 & 7 \\
        \textbf{Citeseer} & 3312 & 4732 & 3312 & 6\\
        \textbf{UAT} & 1190 & 13599 & 239 & 4\\
        \bottomrule
    \end{tabular}
\end{table}

\subsection{Comparative Results}

\begin{figure*}[htbp]
    \begin{subfigure}[b]{\linewidth}
        \centering
        \begin{subfigure}[b]{0.24\linewidth}
            \centering
            \includegraphics[width=\linewidth]{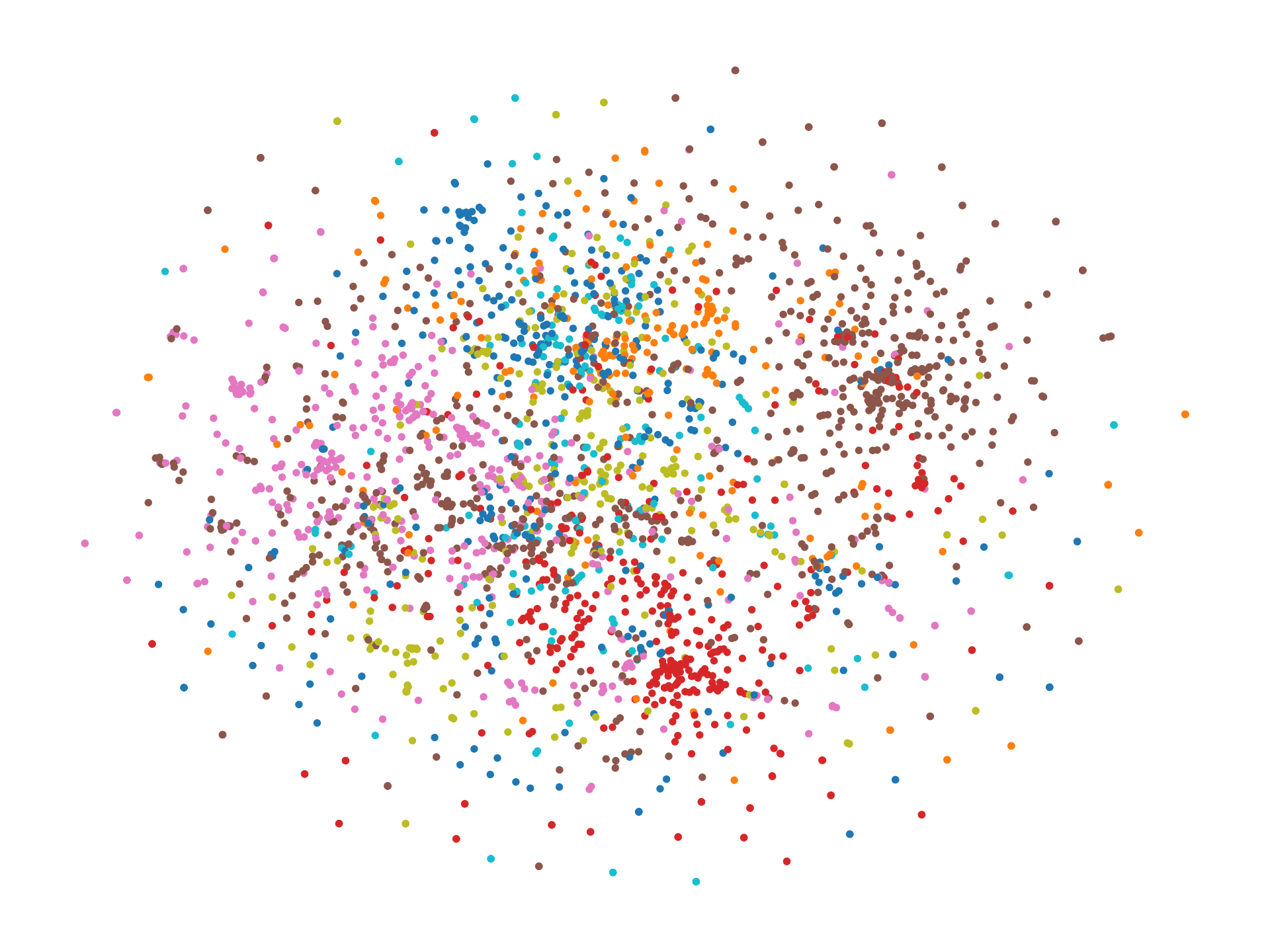}
        \end{subfigure}
        \begin{subfigure}[b]{0.24\linewidth}
            \centering
            \includegraphics[width=\linewidth]{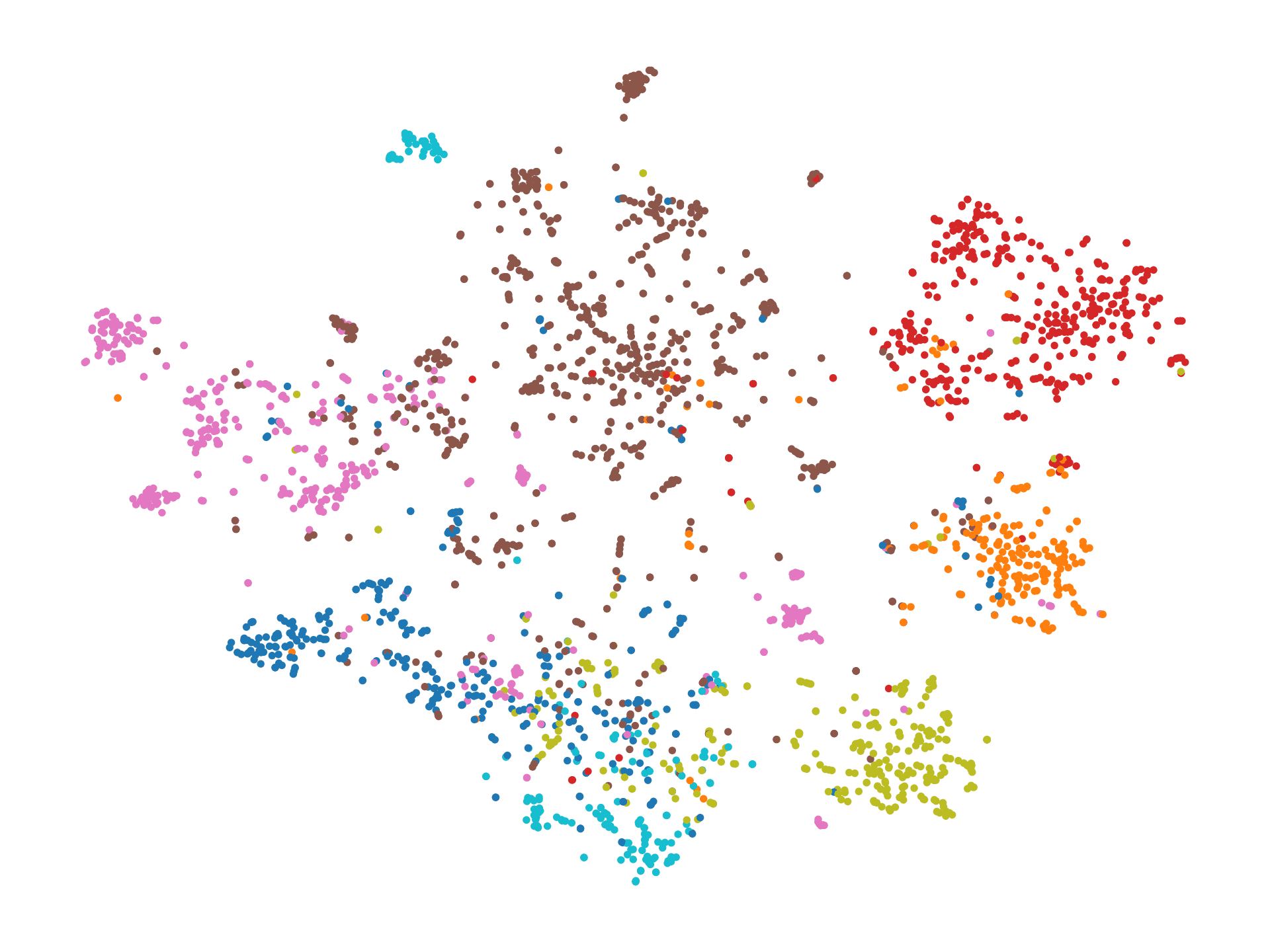}
        \end{subfigure}
        \begin{subfigure}[b]{0.24\linewidth}
            \centering
            \includegraphics[width=\linewidth]{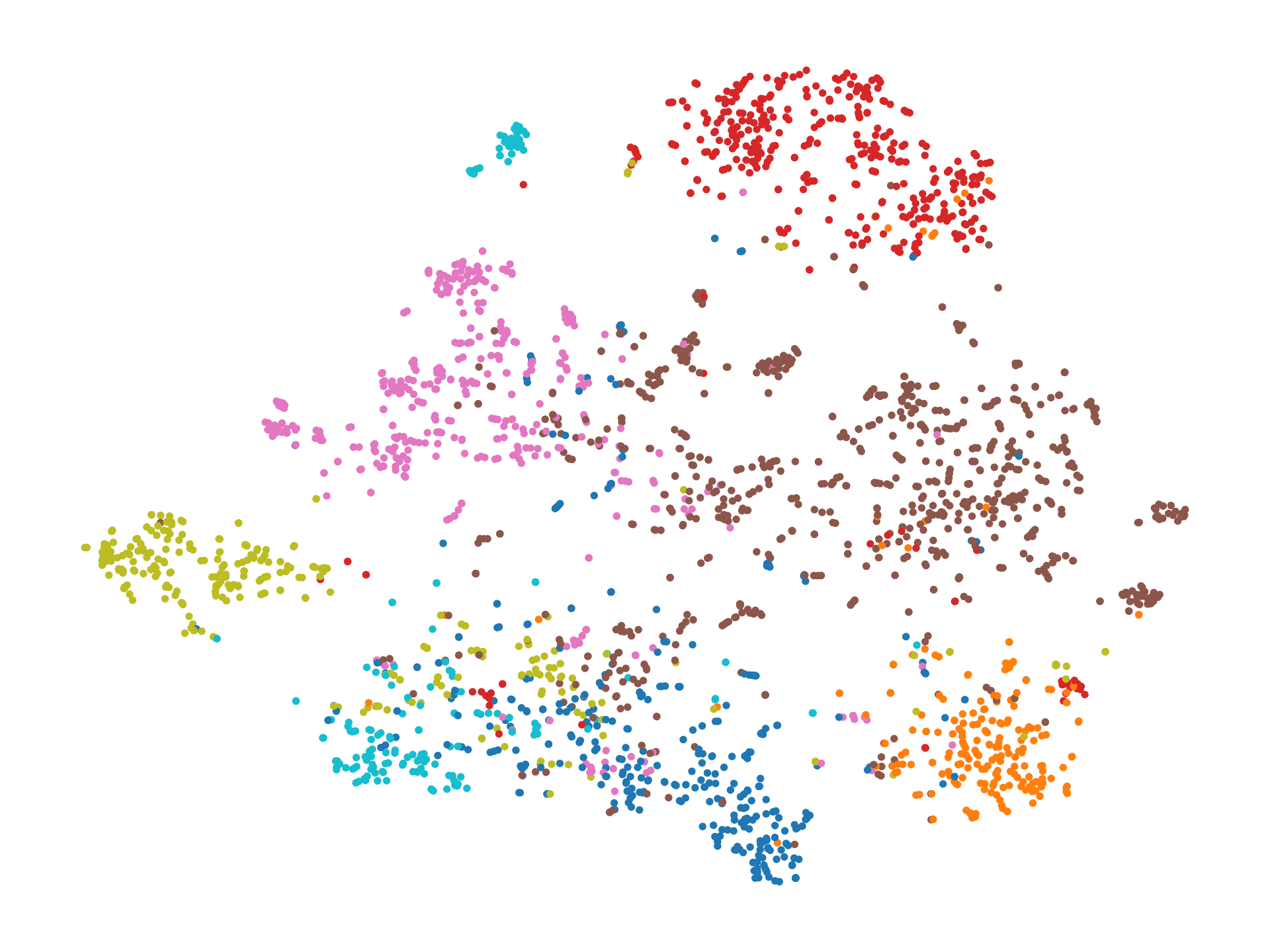}
        \end{subfigure}
        \begin{subfigure}[b]{0.24\linewidth}
            \centering
            \includegraphics[width=\linewidth]{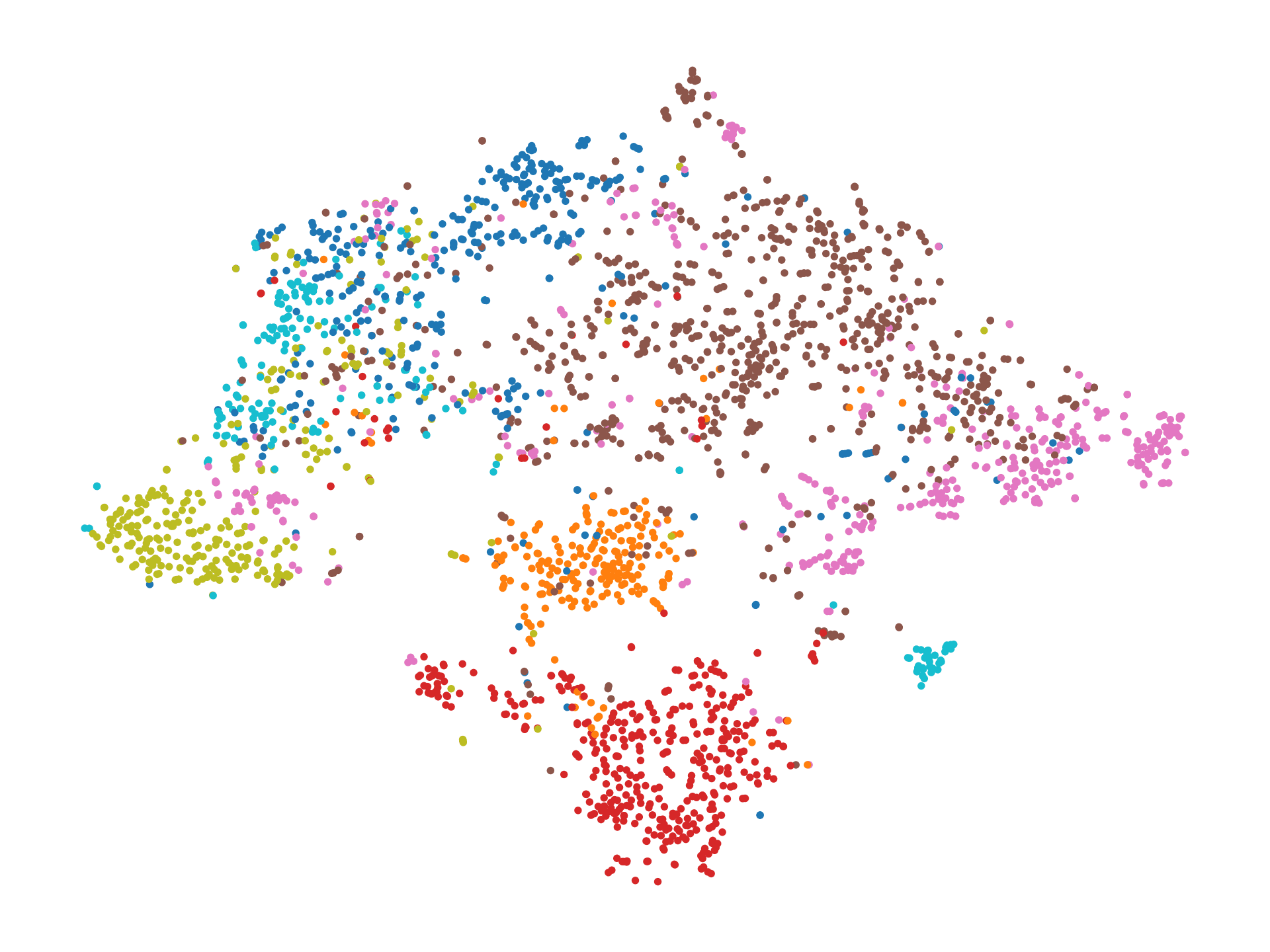}
        \end{subfigure}
        \caption{Cora dataset}
    \end{subfigure}
    
    \begin{subfigure}[b]{\linewidth}
        \centering
        \begin{subfigure}[b]{0.24\linewidth}
            \centering
            \includegraphics[width=\linewidth]{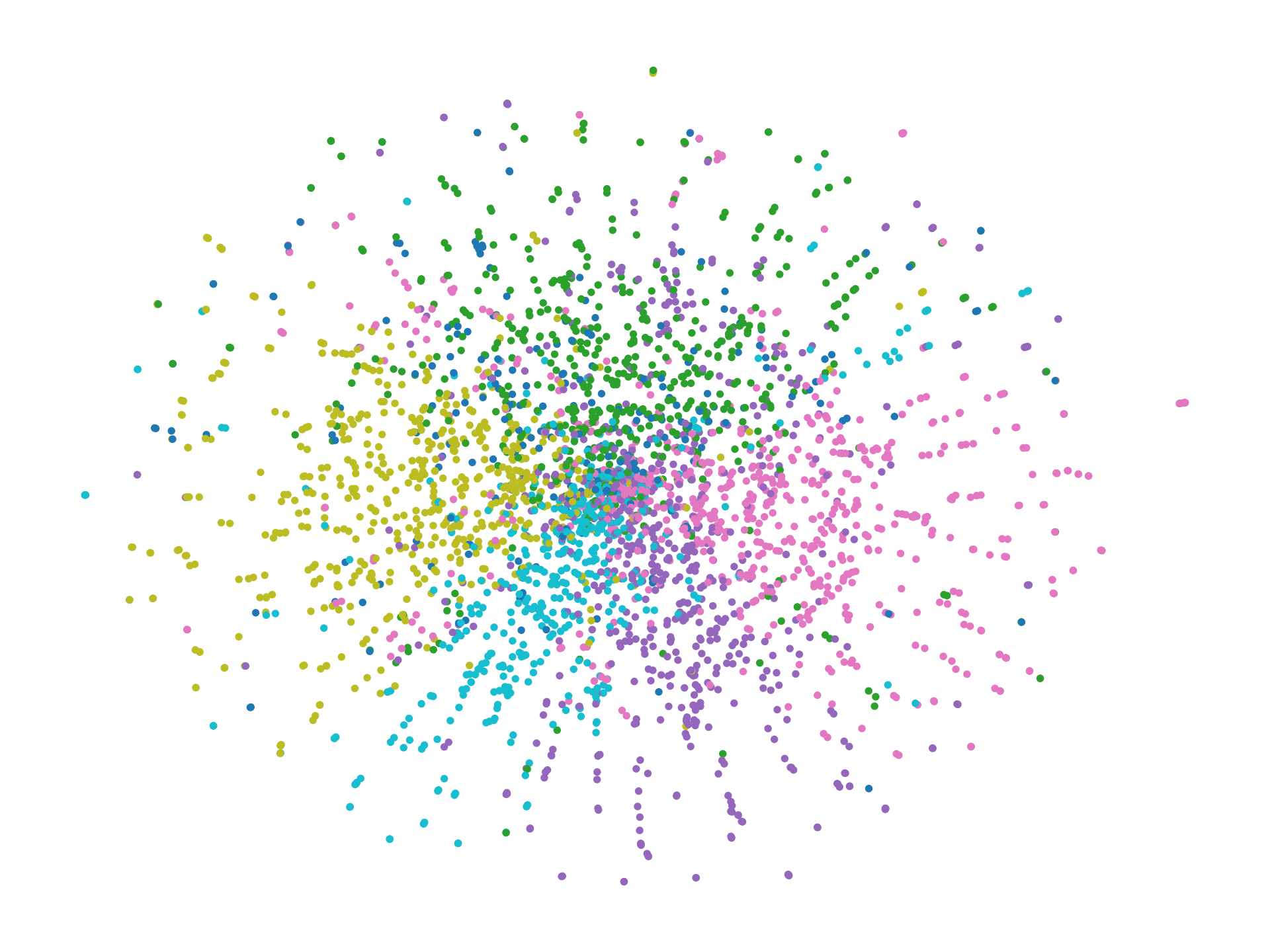}
        \end{subfigure}
        \begin{subfigure}[b]{0.24\linewidth}
            \centering
            \includegraphics[width=\linewidth]{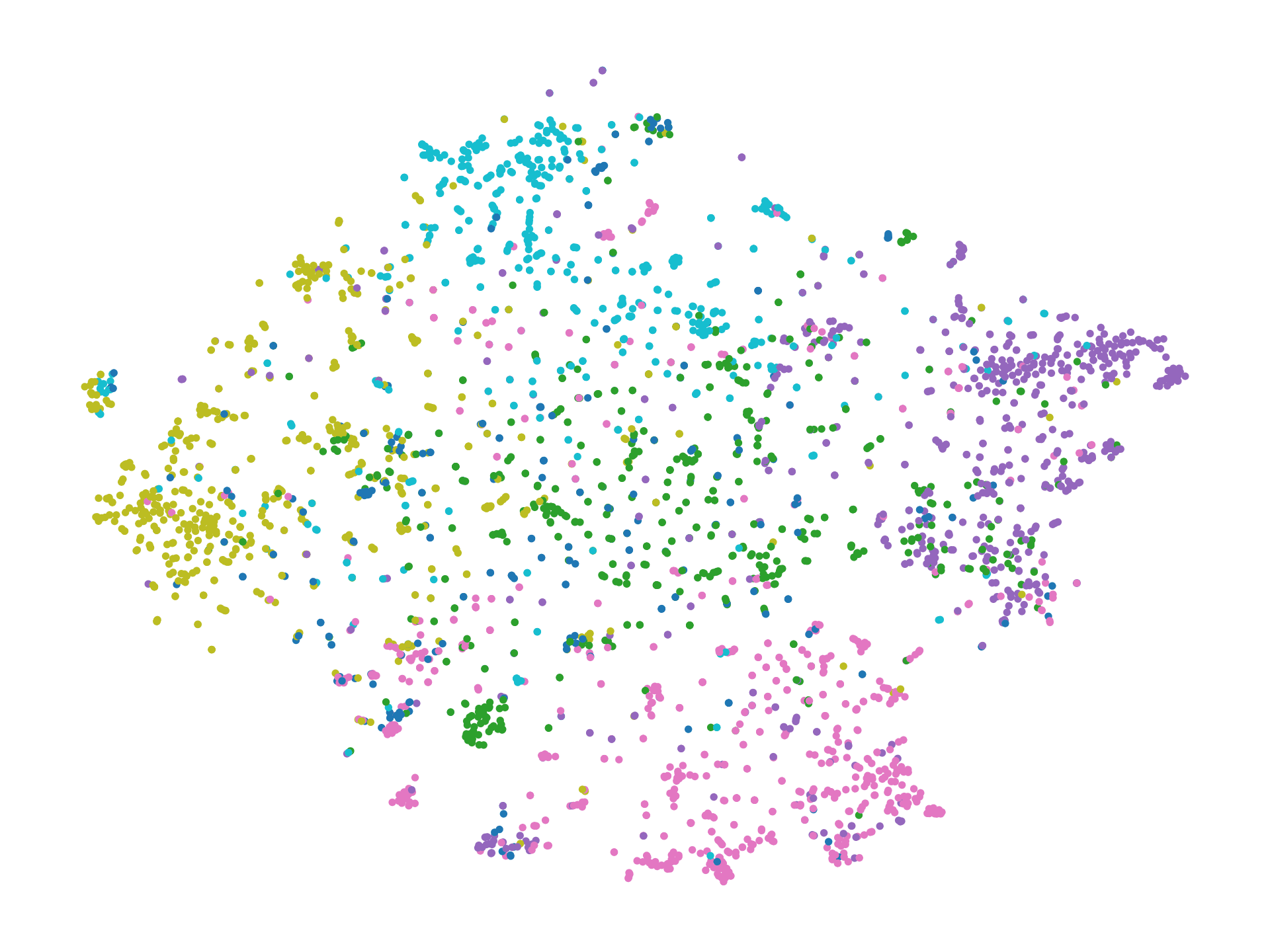}
        \end{subfigure}
        \begin{subfigure}[b]{0.24\linewidth}
            \centering
            \includegraphics[width=\linewidth]{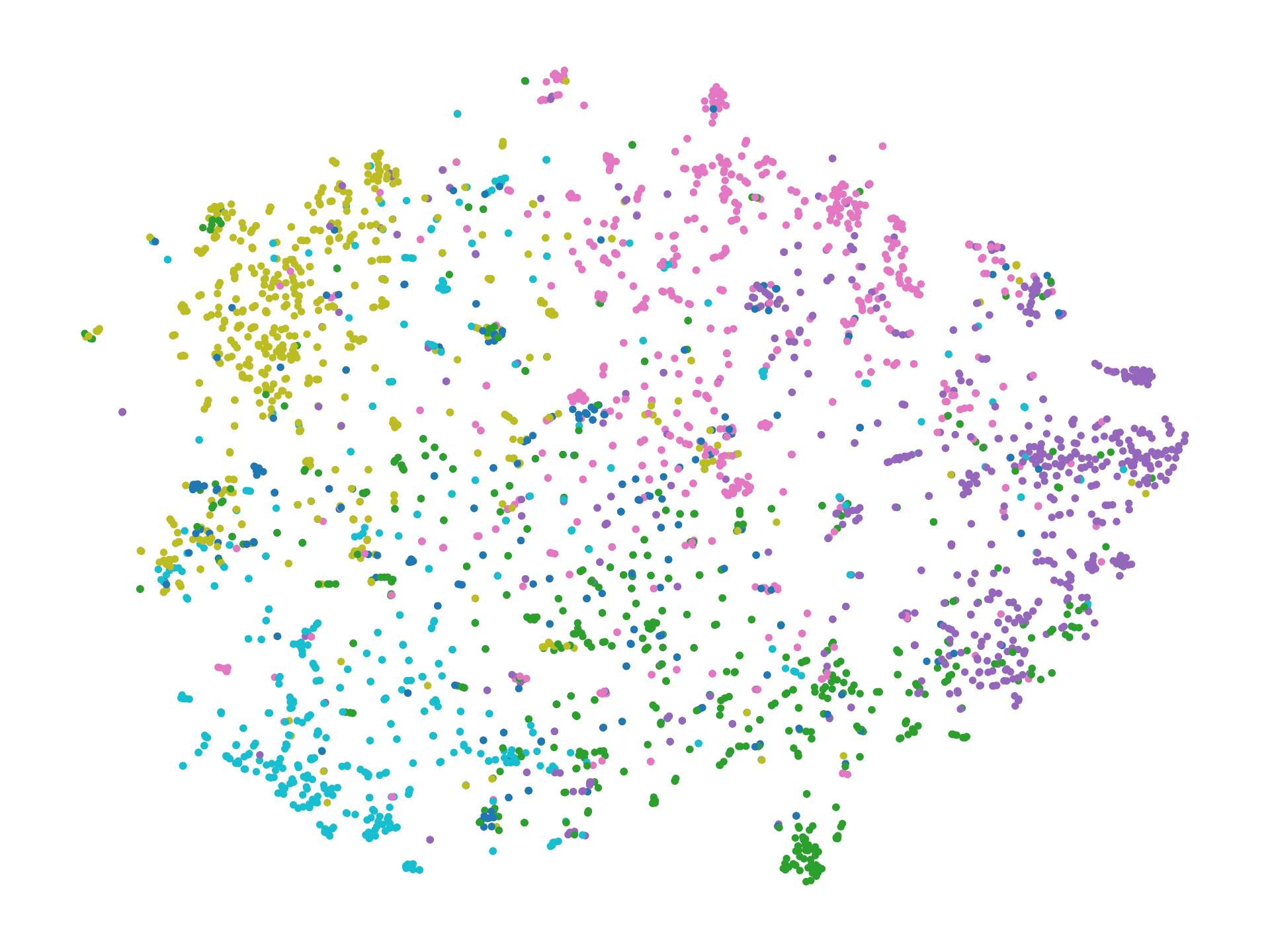}
        \end{subfigure}
        \begin{subfigure}[b]{0.24\linewidth}
            \centering
            \includegraphics[width=\linewidth]{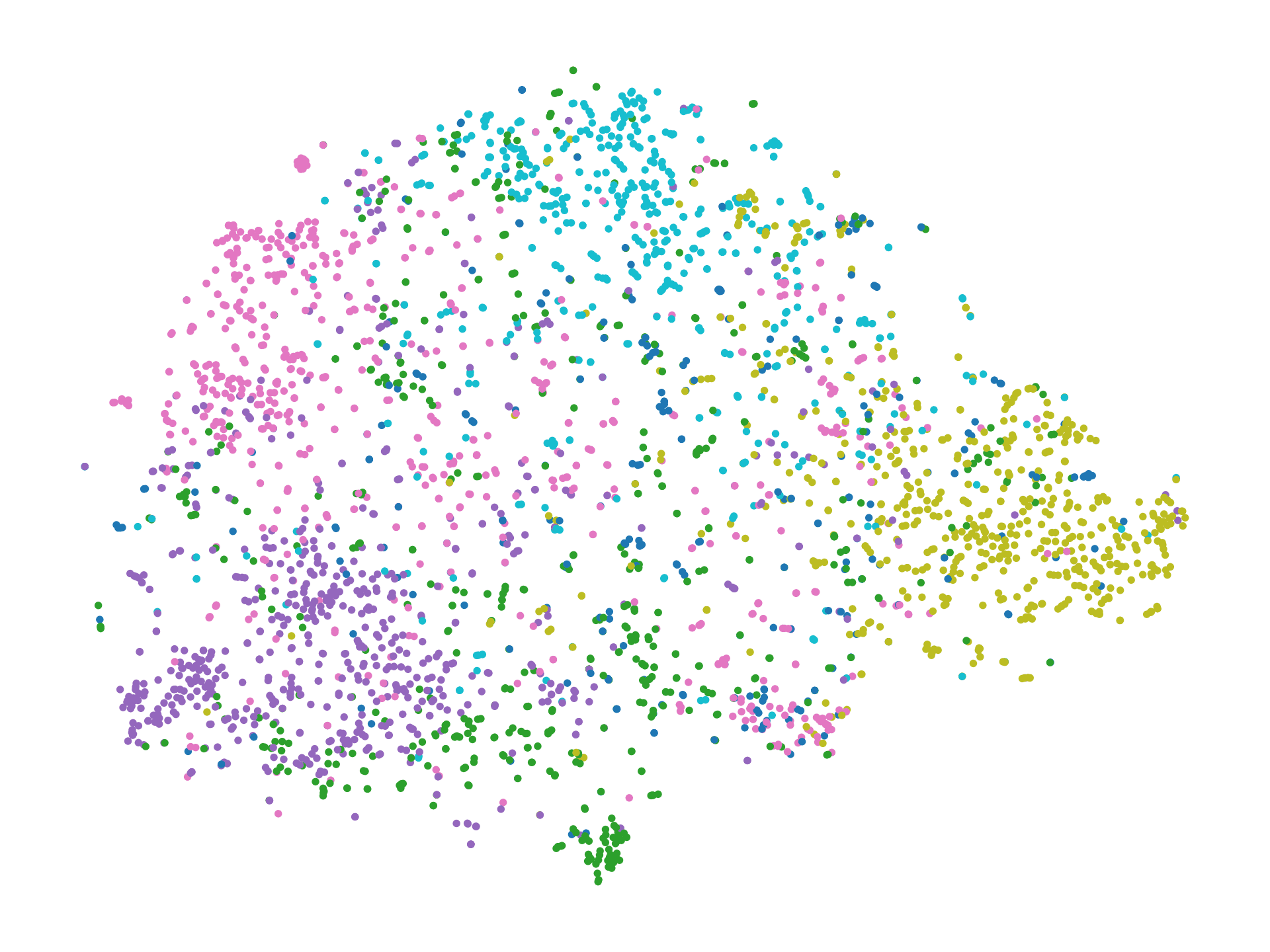}
        \end{subfigure}
        \caption{Citeseer dataset}
    \end{subfigure}
    
    \begin{subfigure}[b]{\linewidth}
        \centering
        \begin{subfigure}[b]{0.24\linewidth}
            \centering
            \includegraphics[width=\linewidth]{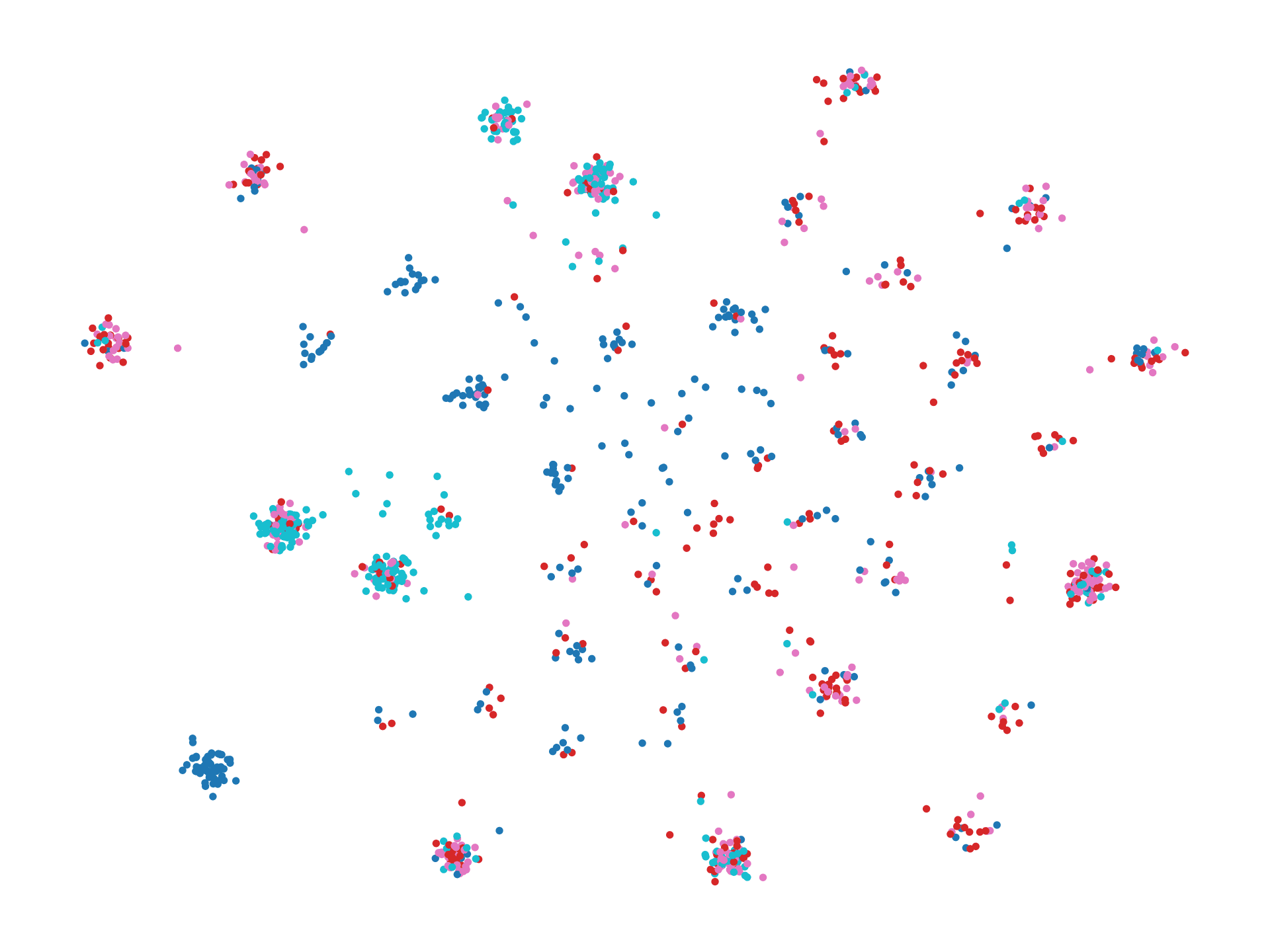}
        \end{subfigure}
        \begin{subfigure}[b]{0.24\linewidth}
            \centering
            \includegraphics[width=\linewidth]{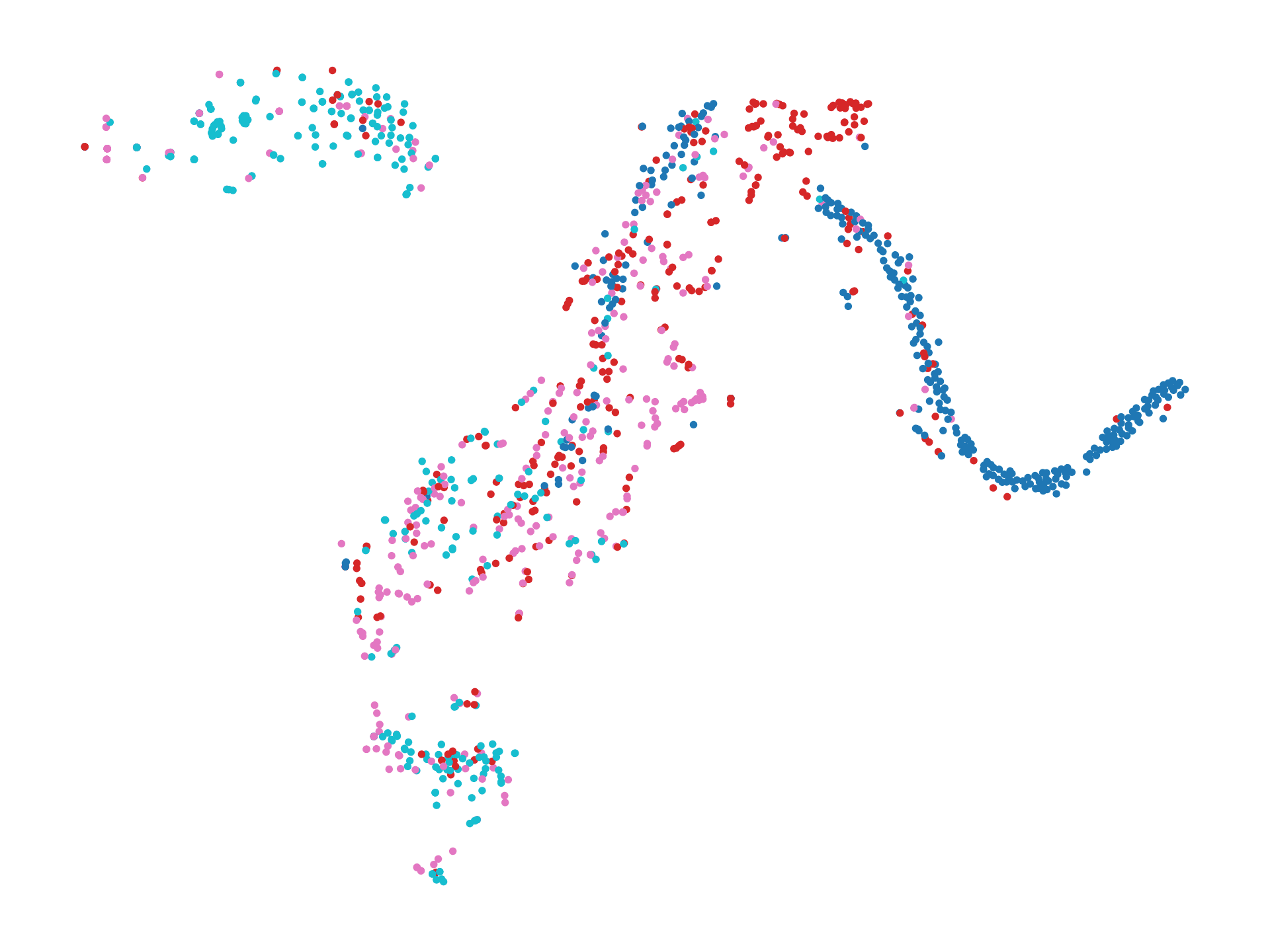}
        \end{subfigure}
        \begin{subfigure}[b]{0.24\linewidth}
            \centering
            \includegraphics[width=\linewidth]{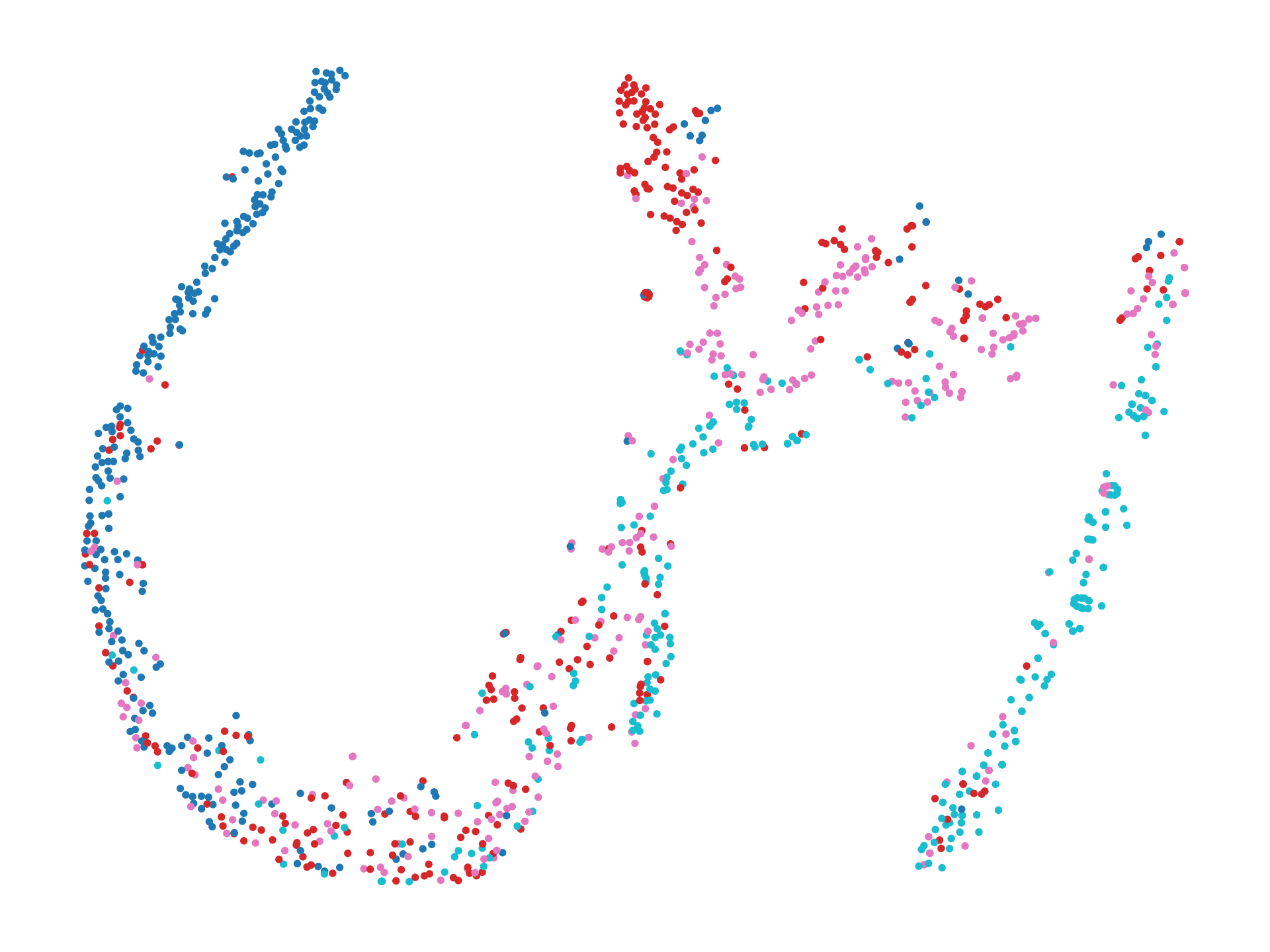}
        \end{subfigure}
        \begin{subfigure}[b]{0.24\linewidth}
            \centering
            \includegraphics[width=\linewidth]{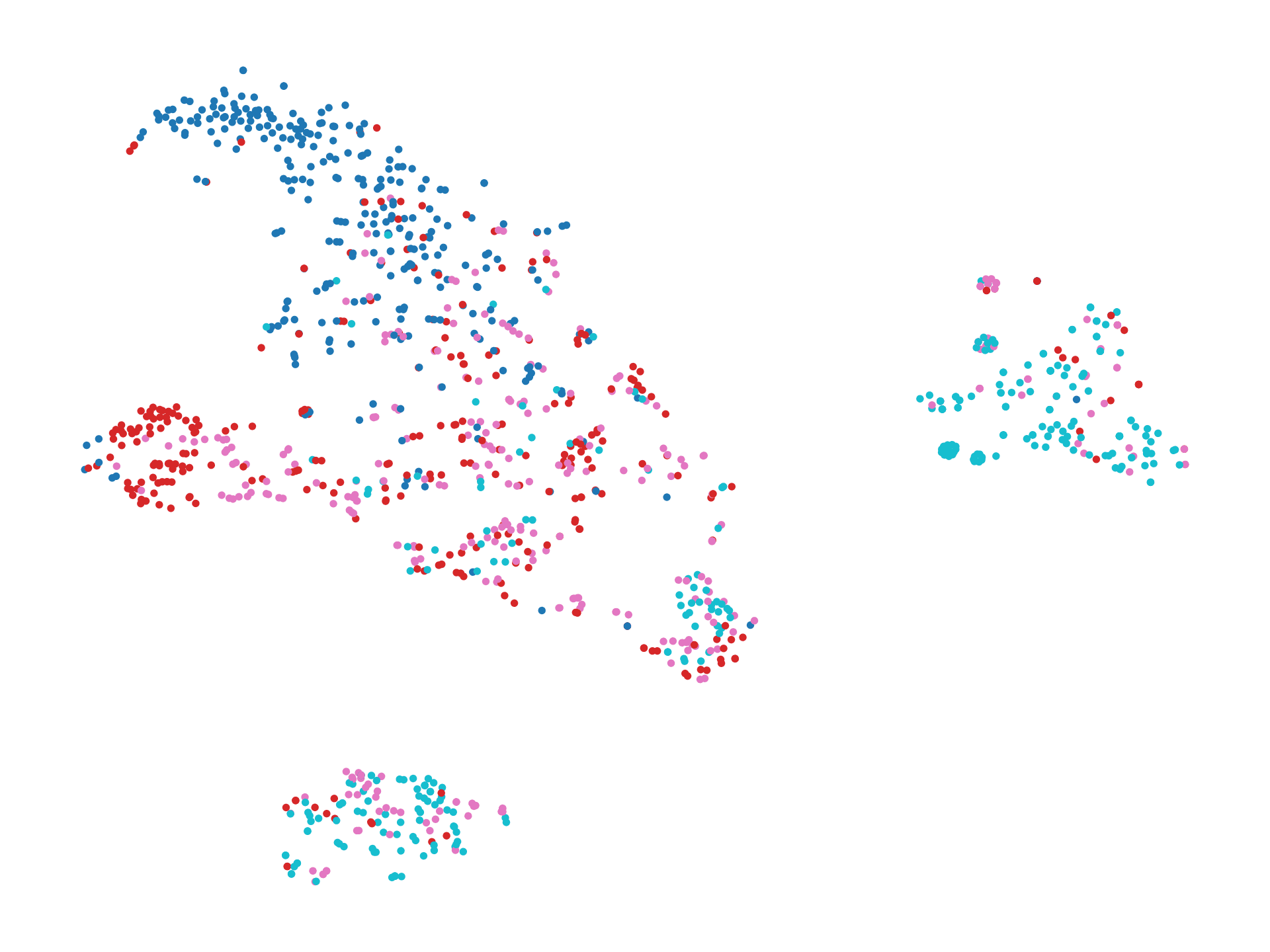}
        \end{subfigure}
        \caption{UAT dataset}
    \end{subfigure}
    \caption{Latent spaces visualization with t-SNE using: (1) Raw features, (2) GAE encoded features, (3) ARGA encoded features, (4) MVGRL encoded features. Each color depicts a single class.}
    \label{fig:latent_spaces}
\end{figure*}

The results are summarized in Table \ref{tab:expresults}. They indicate that each algorithm has its strengths. The Markov algorithm \cite{Van_Dongen_2008} excels in terms of accuracy on two out of the three graphs, achieving scores of up to $89.8\%$. The Leiden algorithm \cite{leiden} optimizes modularity more effectively, as it is designed to optimize this specific metric. Additionally, Deep Graph Clustering methods achieve relatively high NMI (Normalized Mutual Information) and ARI (Adjusted Rand Index) scores. Among the three learning approaches (reconstructive, adversarial, and contrastive), MVRGL \cite{Hassani_Khasahmadi_2020} generally shows small but notable improvements for two out of three datasets.
Finally, these comparative results show that what can be considered the best method depends on the objective of the task, with each algorithm being better at optimizing a particular metric.

\subsection{Latent Spaces Visualization}

Deep Graph Clustering models primarily rely on the transformation of the data into a latent space. Figure \ref{fig:latent_spaces} illustrates the visualization of the latent space for the three models for each data set using the t-SNE algorithm \cite{tsne}.

\section{Applications} \label{sec:applications}
Graph Clustering has experienced significant growth over the past decades, finding numerous real-world applications. The advent of Deep Learning methods has expanded the scope of these applications.

In social network analysis, Graph Clustering is used for community detection \cite{Karatas_Sahin_2018, Mishra_Schreiber_Stanton_Tarjan_2007}. In recommendation systems, it helps group user interests \cite{Jang_Kim_Ha}.

In medical science and bioinformatics, Graph Clustering is applied to biological sequence analysis \cite{Smirnov_Warnow_2021} and the analysis of medical big data \cite{Jiang_Han_Yu_Ding_2023}.

In addition, graph clustering is utilized in computer vision for tasks such as image segmentation \cite{Gammoudi_Mahjoub_Guerdelli_Gammoudi_Mahjoub_Guerdelli_2020,Aflalo_2023_ICCV}, video analysis \cite{Alwassel_Mahajan_Korbar_Torresani_Ghanem_Tran_2020}, and face analysis \cite{Wang_Zheng_Li_Wang_2019}.

\section{Conclusion} \label{sec:conclusion}

In this paper, we explore advanced graph clustering techniques, ranging from traditional methods, such as Spectral Clustering and Markov Clustering, to Deep Graph Clustering methods, in three different paradigms. We have shown in experimental results that each method has its strengths and weaknesses, and the chosen method for a task depends on the objective.

However, several limitations remain in this field of research. One of the primary challenges is the scalability of these methods to very large graphs. Although our experiments were conducted on standard benchmark datasets, real-world graphs often contain millions of nodes and edges, posing significant computational challenges. Additionally, the choice of hyperparameters -- mainly in Deep Graph Clustering methods -- is crucial and quite a challenging task.

Future research should focus on addressing the scalability issue by developing more efficient algorithms and maybe leveraging parallel computing. Another promising direction is the integration of multi-view data, where different types of relationships between nodes can be simultaneously considered to improve clustering accuracy. Exploring the use of self-supervised learning for graph clustering could also provide new avenues for enhancing performance without requiring extensive labeled data.

In general, this paper contributes to the growing field of graph clustering by introducing and evaluating advanced methods that leverage the power of deep learning and adversarial techniques. These methods not only improve clustering accuracy but also open up new possibilities for applications in various domains. As research in this area continues to evolve, we anticipate further advancements that will overcome current limitations and expand the potential uses of graph clustering.

\appendix
\section{Best Hyperparameters for Deep Graph Clustering Models}

The table \ref{tab:bestparamsdeep} presents, for reproducibility, the hyperparameters that gave the best evaluations.

\begin{table}[H]
    \centering
    \begin{tabular}{l|ccccc}
        \toprule
        \textbf{Model}& $\alpha$ & $d_h$ & $e$ & $L_{\text{GCN}}$ & $L_{\text{FC}}$ \\
        \midrule
        GAE & 0.001 & 32 & 50 & 2 & $N.A.$ \\
        ARGA & 0.001 & 32 & 50 & 2 & $N.A.$ \\
        MVGRL & 0.001 & 128 & 40 & 2 & 0\\
        \bottomrule
    \end{tabular}
    \caption{Best hyperparameters for Deep Graph Clustering models ($\alpha$ is the learning rate, $d_h$ the latent node dimension, $e$ the number of training epochs, $L_{\text{GCN}}$ the number of GCN layers and $L_{\text{FC}}$ -- if applicable, else $N.A.$ -- the number of fully connected layers).}
    \label{tab:bestparamsdeep}
\end{table}

\addcontentsline{toc}{section}{References}
\printbibliography
\end{multicols}

\end{document}